\documentclass[a4paper,journal]{IEEEtran}
\IEEEoverridecommandlockouts
\usepackage{booktabs}
\usepackage{cite}
\usepackage{floatrow}
\usepackage{microtype}
\usepackage{graphicx}
\usepackage{subfigure}
\usepackage{booktabs}
\usepackage{amsmath,amsthm,amssymb}
\usepackage[ruled,vlined]{algorithm2e}
\usepackage{algorithmic}

\usepackage{multirow}
\usepackage{textgreek}
\usepackage{hyperref}

\usepackage{booktabs}     
\usepackage{amsfonts}     
\usepackage{nicefrac}      
\usepackage{microtype}   
\usepackage{booktabs}
\usepackage[flushleft]{threeparttable}
\usepackage{times}
\usepackage{graphicx} 
\usepackage{lipsum}
\usepackage{mwe}
\usepackage{enumitem}
\usepackage{graphicx,color}
\usepackage{epsfig}
\usepackage{amsmath,amsthm,amssymb}
\usepackage{multirow}
\allowdisplaybreaks
\usepackage{float}
\floatstyle{plaintop}
\restylefloat{table}

\usepackage[nodisplayskipstretch]{setspace}
\usepackage{titlesec}
\titlespacing\section{-2pt}{4pt plus 2pt minus 1pt}{0pt plus 2pt minus 2pt}
\titlespacing\subsection{-4pt}{0pt plus 2pt minus 1pt}{0pt plus 2pt minus 2pt}
\titlespacing\subsubsection{-4pt}{0pt plus 2pt minus 1pt}{0pt plus 2pt minus 2pt}
\newtheoremstyle{exampstyle}
  {1pt} 
  {1pt} 
  {} 
  {} 
  {\bfseries} 
  {.} 
  {.5em} 
  {} 
\theoremstyle{exampstyle}

\newtheorem{thm}{Theorem}
\newtheorem{assumption}{Assumption}

\newtheorem{lem}{Lemma}

\newtheorem{defn}{Definition}

\theoremstyle{remark}
\newtheorem{rem}{Remark}

\renewcommand{\algorithmicrequire}{\textbf{Input:}} 
\renewcommand{\algorithmicensure}{\textbf{Output:}} 
\begin{document}
\title{Asynchronous Training Schemes in Distributed Learning with Time Delay}

\author{Haoxiang Wang, Zhanhong Jiang, Chao Liu, Soumik Sarkar, Dongxiang Jiang \emph{and} Young M. Lee
\thanks{H. Wang is with the Institute for Interdisciplinary Information Sciences (IIIS), Tsinghua University, Beijing, 100084, P.R. China. 

Chao Liu and  Dongxiang Jiang are affiliated with the Department of Energy and Power Engineering, Tsinghua University, Beijing 100084, China. 

Zhanhong Jiang and Young M. Lee are with Johnson Controls, 507 East Michigan St, Milwaukee, WI 53211, USA. 

Soumik Sarkar is with the Department of Mechanical Engineering, Iowa State University, Ames, IA 50011, USA. 

Corresponding author: Chao Liu.}}
\maketitle

\begin{abstract}
In the context of distributed deep learning, the issue of stale weights or gradients could result in poor algorithmic performance. This issue is usually tackled by delay tolerant algorithms with some mild assumptions on the objective functions and step sizes. In this paper, we propose a different approach to develop a new algorithm, called \textbf{P}redicting \textbf{C}lipping \textbf{A}synchronous \textbf{S}tochastic \textbf{G}radient \textbf{D}escent (aka, PC-ASGD). Specifically, PC-ASGD has two steps - the \textit{predicting step} leverages the gradient prediction using Taylor expansion to reduce the staleness of the outdated weights while the \textit{clipping step} selectively drops the outdated weights to alleviate their negative effects. A tradeoff parameter is introduced to balance the effects between these two steps. Theoretically, we present the convergence rate considering the effects of delay of the proposed algorithm with constant step size when the smooth objective functions are weakly strongly-convex and nonconvex. One practical variant of PC-ASGD is also proposed by adopting a condition to help with the determination of the tradeoff parameter. For empirical validation, we demonstrate the performance of the algorithm with two deep neural network architectures on two benchmark datasets. 
\end{abstract}

\begin{IEEEkeywords}
Distributed deep learning; gradient prediction; asynchronous SGD; convergence; time delay
\end{IEEEkeywords}

\section{Introduction}\label{introduction}
The availability of large data sets and powerful computing led to the emergence of deep learning that is revolutionizing many application sectors from the internet industry and healthcare to transportation and energy~\cite{gijzen2013big,wiedemann2019compact}. As the applications are scaling up, the learning process of large deep learning models is looking to leverage emerging resources such as edge computing and distributed data centers 
privacy preserving. In this regard, distributed deep learning algorithms are being explored by the community that leverage synchronous and asynchronous computations with multiple computing agents that exchange information over communication networks~\cite{lian2017asynchronous}. We consider an example setting involving an industrial IoT framework where the data is geographically distributed as well as the computing resources. While the computing resources within a local cluster can operate in a (loosely) synchronous manner, multiple (geographically distributed) clusters may need to operate in an asynchronous manner. Furthermore, communications among the computing resources may not be reliable and prone to delay and loss of information. 

Among various distributed deep learning algorithms, Federated Averaging and its variants are considered to be the state-of-the-art for training deep learning models with data distributed among the edge computing resources such as smart phones and idle computers \cite{hard2018federated,sattler2019robust}. 
The master-slave and peer-to-peer are two categories of distributed learning architectures.
Along with Federated Averaging, variants such as PySyft~\cite{ryffel2018a} and its robust version \cite{deng2020distributionally}, the scalable distributed DNN training algorithms~\cite{strom2015scalable} and more recent distributed SVRG \cite{cen2020convergence} and clustered FL \cite{9174890} are examples of the master-slave architecture. On the other hand, examples of the peer-to-peer architecture include the gossip algorithms~\cite{blot2016gossip,even2020asynchrony,li2021consensus}, and the collaborative learning frameworks~\cite{jiang2017collaborative,liu2019distributed}. 

However, as mentioned earlier, communication delay remains a critical challenge for achieving convergence in an asynchronous learning setting~\cite{chen2016revisiting,tsianos2012communication} and influences the performances of the frameworks above. Furthermore, the amount of delay could be varying widely due to artifacts of wireless communication and different devices. To tackle the influence of varying delays on the convergence characteristics of distributed learning algorithms, this work proposes a novel algorithm, called \textbf{P}redicting \textbf{C}lipping \textbf{A}synchronous \textbf{S}tochastic \textbf{G}radient \textbf{D}escent (aka, PC-ASGD). The goal is to solve the distributed learning problems involving multiple computing or edge devices such as GPUs and CPUs with varying communication delays among them. Different from traditional distributed learning scenarios where synchronous and asynchronous algorithms are considered separately, we take both into account together in a networked setting.

\textbf{Related work}. 
In the early works on distributed learning with master-slave architecture, Asynchronous Stochastic Gradient Descent (ASGD) algorithm has been adopted \cite{dean2012large}, where each local worker continues its training process right after its gradient is added to the global model. The algorithm could tolerate the delay in communication. Later works~\cite{agarwal2011distributed,feyzmahdavian2015an,recht2011hogwild:,zhuang2021fully}
extend ASGD to more realistic scenarios and implement the algorithms with a central server and other parallel workers. Typically, since asynchronous algorithms suffer from stale gradients, researchers have proposed algorithms such as DC-ASGD~\cite{zheng2017asynchronous}, adopting the concept of delay compensation to reduce the influence of staleness and improve the performance of ASGD. For the distributed learning with peer-to-peer architecture, \cite{lian2017asynchronous} proposes the algorithm AD-PSGD (decentralized ASGD algorithm) that deals with the problem of the stale parameter exchange, as well as some theoretical analysis for the algorithm performance under bounded
delay. 
\cite{liang2020asynchrounous} also proposes a similar algorithm with some variations in the assumptions. However, these algorithms do not provide empirical or theoretical analysis on the
impacts of delay in detail. Additional works such as using a central agent for control~\cite{nair2017wildfire:}, requiring prolonged communication~\cite{TsianosEfficient}, utilizing stochastic primal-dual method~\cite{Lan2017Communication}, and adopting importance sampling~\cite{du2020asynchronous}, have also been done to address the communication delay in the decentralized setting.
More recently, \cite{rigazzi2019dc-s3gd:} proposes the DC-s3gd algorithm to enable large-scale decentralized neural network training with the consideration of delay. 
\cite{2020arXiv200107704Z}, \cite{venigalla2020adaptive}, \cite{chen2019communication} and \cite{2020arXiv200100112A} also develop algorithms for asynchronous decentralized training for neural networks, while theoretical guarantee is not provided.

\begin{table*}[htbp] 
\caption{Comparisons between asynchronous algorithms}
\centering
\begin{footnotesize}
\label{tabalgcomp}
\begin{tabular}{llllllll}
\hline
Methods  & $f$          & $\nabla f$ & Delay Ass. & Con.Rate  & D.C.  & G.C.                                                  & A.S. \\ \hline
ASGD~\cite{dean2012large}                                 & Non-convex & Lip.                    & Bou.       & $\mathcal{O}(\frac{1}{\sqrt{T}})$ & No & No & No \\ \cline{2-8} 
\multirow{2}{*}{DC-ASGD~\cite{zheng2017asynchronous}}             & Str-con    & Lip.                    & Bou.     & $\mathcal{O}(\frac{1}{T})$                        & No & Yes & No   \\
                                     & Non-convex & Lip.                    & Bou.     & $\mathcal{O}(\frac{1}{\sqrt{T}})$ & No & Yes & No   \\ \cline{2-8} 
AD-PSGD~\cite{lian2017asynchronous}                              & Non-convex & Lip.\&Bou.              & Bou.       & $\mathcal{O}(\frac{1}{\sqrt{T}})$ & Yes & No & No  \\ \cline{2-8} 
DC-s3dg~\cite{rigazzi2019dc-s3gd:}                              & Non-convex & Lip.                    & Unbou.     & N/A                                                       & Yes & Yes & No   \\ \cline{2-8} 
\multirow{2}{*}{PC-ASGD (This paper)} & Weakly Str-con    & Lip.             & Bou.     & $\mathcal{O}(\epsilon^T)$            & Yes & Yes & Yes \\
                                     & Non-convex & Lip.             & Bou.     & $\mathcal{O}(\frac{1}{T})$  & Yes & Yes & Yes   
\\ \hline
\end{tabular}\\
\end{footnotesize}
\begin{tablenotes}
\item \scriptsize{Con.Rate: convergence rate, Str-con: strongly convex. Lip.\& Bou.: Lipschitz continuous and bounded. Delay Ass.: Delay Assumption. Unbou.: Unbounded. $T$: Total iterations. D.C.: decentralized computation. G.C.: Gradient Compensation. A.S.: Alternant Step, $\epsilon \in (0,1)$ is a positive constant. Note that the convergence rate of PC-ASGD is obtained by using the constant step size.}
\end{tablenotes}
\end{table*}

\vspace{5pt}
\textbf{Contributions}. The contributions of this work are specifically as follows. (i) A novel algorithm, called PC-ASGD for distributed learning is proposed to tackle the convergence issues due to the varying communication delays. Built upon ASGD, the PC-ASGD algorithm consists of two steps. While the predicting step leverages the gradient prediction using Taylor expansion to reduce the staleness of the outdated weights, the clipping step selectively drops the outdated weights to alleviate their negative effects. To balance the effects, a tradeoff parameter is introduced to combine these two steps. (ii) We show that with a proper constant step size, PC-ASGD can converge to the neighborhood of the optimal solution at a linear rate for weakly strongly-convex functions while at a sublinear rate for nonconvex functions (specific comparisons with other related existing approaches are listed in Table \ref{tabalgcomp}). \textcolor{black}{We also model the delay and take it into consideration in the convergence analysis.} (iii) PC-ASGD is deployed on distributed GPUs with two datasets CIFAR-10 and CIFAR-100 by using PreResNet110 and DenseNet architectures. The proposed algorithm outperforms the existing delay tolerant algorithm as well as the variants of the proposed algorithm using only the predicting step or the clipping step.

\section{Formulation and Preliminaries}
\label{secPrelim}
Consider $N$ agents in a networked system such that their interactions are driven by a graph $\mathcal{G}$, where $\mathcal{G} = \{V,E\}$, where $V=\{1,2,..,N\}$ indicates the node or agent set, $E\subseteq V\times V$ is the edge set. Throughout the paper, we assume that the graph is undirected and connected. The connection between any two agents $i$ and $j$ can be determined by their physical connections, leading to the communication between them. Traditionally, if agent $j$ is in the neighborhood of agent $i$, they can communicate with each other. Thus, we define the neighborhood for any agent $i$ as 
$
Nb(i):=\{j \in V|(i,j)\in E\;or\,\,j=i\}
$.
Rather than considering synchronization and asynchronization separately, this paper considers both scenarios together by defining the following terminologies. 
\vspace{5pt}
\begin{defn}\label{reiable_neighbor}
At a time step $t$, an agent $j$ is called a \textbf{\textit{reliable neighbor}} of the agent $i$ if agent $i$ has the state information of agent $j$ up to $t-1$.
\end{defn}
\vspace{3pt}
\begin{defn}\label{unreliable_neighbor}
At a time step $t$, an agent $j$ is called an \textbf{\textit{unreliable neighbor}} of the agent $i$ if agent $i$ has the state information of agent $j$ only up to $t-\tau$, where $\tau$ is the so-called \textit{delay} and $1 < \tau < \infty$.
\end{defn}
Definitions~\ref{reiable_neighbor} and~\ref{unreliable_neighbor} allow us to perceive the delay problem in the decentralized learning with a new perspective that depends on the amount of delay. One agent can selectively make use of the outdated information from unreliable neighbors or completely drop such information. The first scenario is related to most previous works on asynchronous delay tolerant approaches as it involves a gradient prediction technique to reduce the negative effects of stale parameters. The second scenario corresponds to most synchronous schemes since the agent only collects information from the reliable neighbors. Thus, inside the neighborhood of an agent, there are reliable and unreliable neighbors respectively and this work aims at studying how to effectively tackle issues such as negative impacts that delays may bring on the performance. For analysis, we define a set for reliable neighbors of agent $i$ as:
$\mathcal{R}:=\{j\in Nb(i)\;|\;p(x^j=x^j_{t-1}|t)=1\}$, where $p(x^j=x^j_{t-1}|t)=1$ is the probability, implying that agent $j$ has the state information $x$ up to the time $t-1$, i.e., $x^j_{t-1}$. Then we can have the set for unreliable neighbors such that $\mathcal{R}^c=Nb\setminus \mathcal{R}$.

\textcolor{black}{Note that the delay varies in the asynchronous learning scheme, and there are two types of asynchronization, (i) fixed value of delays  \cite{zheng2017asynchronous,rigazzi2019dc-s3gd:} and (ii) time-varying delays \cite{dean2012large,lian2017asynchronous} along the learning process. We follow the first setting in this work to implement the experiments. The definition and analysis can also be applicable for the latter case when the delay $\tau$ changes to a time-varying vector.
}

Consider the decentralized empirical risk minimization problems, which can be expressed as the summation of all local losses incurred by each agent:
\begin{equation}
\textnormal{min}\,\,\ F(\mathbf{x}):=\sum_{i=1}^{N}\sum_{s \in \mathcal{D}_{i}}f_{i}^{s}(x)
\end{equation}
where $\mathbf{x}=[x^1;x^2;...;x^N]$, $x^i$ is the local copy of $x\in\mathbb{R}^d$, $\mathcal{D}_i$ is a local data set uniquely known by agent $i$, $f^s_i:\mathbb{R}^d\to\mathbb{R}$ is the incurred local loss of agent $i$ given a sample $s$. 
Based on the above formulation, we then assume everywhere that our objective function is bounded from below and denote the minimum by $F^{*}:=F(\mathbf{x}^{*})$ where $\mathbf{x}^{*}:= \textnormal{argmin} \;F(\mathbf{x})$. Hence $F^{*}>-\infty$. Moreover, all vector norms refer to the Euclidean norm while matrix norms refer to the Frobenius norm. Some necessary definitions and assumptions are given below for characterizing the main results.

\vspace{5pt}
\begin{assumption}\label{assump1}
Each objective function $f_{i}$ is assumed to satisfy the following conditions: a) $f_{i}$ is $\gamma_{i}-smooth$; b) $f_{i}$ is proper (not everywhere infinite) and coercive.
\end{assumption}
\vspace{3pt}
\begin{assumption}\label{assump2}
A mixing matrix $\underline{W}\in\mathbb{R}^{N\times N}$ satisfies a) $\mathbf{1}^\top \underline{W}=\mathbf{1}^\top,\underline{W}\mathbf{1}^\top=\mathbf{1}^\top$; b) $\text{Null}\{I-\underline{W}\}=\text{Span}\{\mathbf{1}\}$; c) $I\succeq \underline{W} \succ 0$.
\end{assumption}
\vspace{3pt}
\begin{assumption}\label{assump3}
The stochastic gradient of $F$ at any $\mathbf{x}$ is denoted by $\mathbf{g}(\mathbf{x})$, such that
a) $\mathbf{g}(\mathbf{x})$ is the unbiased estimate of gradient $\nabla F(\mathbf{x})$; b) The variance is uniformly bounded by $\sigma^2$, i.e.,$\mathbb{E}[\|\mathbf{g}(\mathbf{x})-\nabla F(\mathbf{x})\|^{2}]\leq \sigma^{2}$; c) The second moment of $\mathbf{g}(\mathbf{x})$ is bounded, i.e., $\mathbb{E}[\|\mathbf{g}(\mathbf{x})\|^{2}]\leq G^{2}$.
\end{assumption}
Given Assumption~\ref{assump1}, one immediate consequence is that $F$ is $\gamma_m:=\textnormal{max}\{\gamma_1,\gamma_2,...,\gamma_N\}$-smooth at all $\mathbf{x}\in\mathbb{R}^{dN}$. 
The main outcome of Assumption~\ref{assump2} is that the mixing matrix $\underline{W}$ is doubly stochastic matrix and that we have  
$e_{1}(\underline{W})=1>e_{2}(\underline{W})\geq..\geq e_{N}(\underline{W})>0$, where $e_{z}(\underline{W})$ denotes the $z$-th largest eigenvalue of $\underline{W}$.
In Assumption~\ref{assump3}, the first two are quite generic. While the third part is much weaker than the bounded gradient that is not necessarily applicable to quadratic-like objectives.

\section{PC-ASGD}
\label{secMethod}
\subsection{Algorithm Design}
We present the specific update law for our proposed method, PC-ASGD in Algorithm~\ref{pcasgd}. 
In Algorithm~\ref{pcasgd}, for the predicting step (line 6), any agent $k$ that is unreliable has delay when communicating its weights with agent $i$. To compensate for the delay, we adopt the Taylor expansion to approximate the gradient for each time step. The predicting gradient (or delay compensated gradient) is denoted by $g^{dc}_k(x^k_{t-\tau})$, which is expressed as follows
\begin{equation}\label{delay_comp_gradient}
\begin{split}
g^{dc,r}_k(x^k_{t-\tau}) 
&= \sum^{\tau-1}_{r=0} g_k(x^k_{t-\tau}) \\&+\lambda g_k(x^k_{t-\tau}) \odot g_k(x^k_{t-\tau})
 \odot (x^i_{t-\tau+r} - x^i_{t-\tau}),
\end{split}
\end{equation}
where $\lambda$ is a positive constant in $(0,1]$ and the term $\lambda g_k(x^k_{t-\tau}) \odot g_k(x^k_{t-\tau})$ is an estimate of the Hessian matrix, $\nabla g_k(x^k_{t-\tau})$. Due to the limit of space, we omit the details of deriving Eq.~\ref{delay_comp_gradient}, referring interested readers to the Appendix. We define another doubly stochastic matrix $\tilde{\underline{W}}\in\mathbb{R}^{N\times N}$ that follows Assumption~\ref{assump2} for the clipping step. 

 \vspace{5pt}
\renewcommand{\algorithmicrequire}{\textbf{Input:}}
 \renewcommand{\algorithmicensure}{\textbf{Output:}}
 \begin{algorithm}[htb]  
 \caption{PC-ASGD}
 \label{pcasgd}
 \begin{algorithmic}[1] 
 \REQUIRE number of agents \(N\), learning rate $\eta > 0$,  agent interaction matrices \(\underline{W}\), \(\tilde{\underline{W}}\), number of epochs $T$, the tradeoff parameter $0\leq\theta_t\leq 1, t\in\{0,1,...,T-1\}$
 \ENSURE the models' parameters in agents $x_{T}^i$,$i=1,2,...N$
 \STATE \textbf{Initialize} all the agents' parameters $x_{0}^i$, $i=1,2,...N$
 \STATE Do broadcast to identify the clusters of reliable agents and the delay $\tau$
 \STATE $t=0$
 \WHILE{$epoch\;t < T$} 
 \FOR{each agent $i$}
 \STATE Predicting Step: $x^i_{t+1, pre} = \sum_{j\in \mathcal{R}}w_{ij}x^j_t - \eta g_i(x^i_t)+\sum_{k\in \mathcal{R}^c}w_{ik}(x^k_{t-\tau} - \eta g^{dc}_k(x^k_{t-\tau}))$
 \STATE Clipping Step: $x^i_{t+1, cli} = \sum_{j\in \mathcal{R}}\tilde{w}_{ij}x^j_t - \eta g_i(x^i_t)$
 \STATE $x^i_{t+1}=\theta_tx^i_{t+1,pre}+(1-\theta_t)x^i_{t+1,cli}$
 \ENDFOR
 \STATE $t=t+1$
 \ENDWHILE
 \end{algorithmic}
 \end{algorithm}

Different from the DC-ASGD, which significantly relies on a central server to receive information from each agent, our work removes the dependence on the central server, and instead constructs a graph for all of agents.
The clipping step (line 7) essentially rejects information from all the unreliable neighbor in the neighborhood of one agent. One observation can be made from the predicting and clipping steps is that the weights for consensus terms are different. In this context, for the clipping step, the weight values of some edges associated with the underlying static graph has been changed accordingly, but the connections of the graph still keep fixed. Subsequently, the equality in line 8 balances the tradeoff between the predicting and clipping steps. In practice, the determination of $\theta_t$ results in some practical variants. In the empirical study presented in Section 5, one can see that $\theta_t$ is either 0 or 1 by leveraging one condition, which implies that in each epoch, only one step is adopted. Additionally, $\theta_t$ can be fixed as 0 or 1, yielding two other variants shown in the experiments, C-ASGD or P-ASGD. However, for the sake of generalization, we provide the analysis for the combined steps (line 8). Before concluding this section, we give the compact form of the combination of PC steps and defer the detailed analysis to the Appendix.

Since the term $\sum_{k \in \mathcal{R}^c}w_{ik}\sum_{r=0}^{\tau-1}g_k^{dc,r}(x_{t}^{k})$ applies to unreliable neighbors only, for the convenience of analysis, we expand it to the whole graph. It means that we establish an expanded graph to cover all of agents by setting some elements in the mixing matrix $\underline{W}'\in\mathbb{R}^{N\times N}$ equal to $0$, but keeping the same connections as in $\underline{W}$. By setting the current time as $t+\tau$, the compact form in line 8 can be rewritten as:
\begin{equation}\label{g_pcasgd1}
\begin{split}
    \mathbf{x}_{t+\tau+1}=\mathcal{W}_{t+\tau}\mathbf{x}_{t+\tau}-\eta(\mathbf{g}(\mathbf{x}_{t+\tau})+\theta_{t+\tau}\sum_{r=0}^{\tau-1}W'\mathbf{g}^{dc,r}(\mathbf{x}_{t}))
\end{split}
\end{equation}
$\mathcal{W}_{t+\tau}$ is denoted by $\theta_{t+\tau}W+(1-\theta_{t+\tau})\tilde{W}$, where $W=\underline{W}\otimes I_{d\times d}$, $\tilde{W} = \tilde{\underline{W}}\otimes I_{d\times d}$, and $W'=\underline{W}'\otimes I_{d\times d}$. Though the original graphs corresponding to the predicting and clipping steps are static, the equivalent graph $\mathcal{W}_{t+\tau}$ has become time-varying due to the time-varying $\theta$ value.
\section{Convergence Analysis}
\label{secMainresults}
This section presents convergence results for the PC-ASGD. We show the consensus estimate and the optimality for both weakly strongly-convex and nonconvex smooth objectives.
The consensus among agents (aka, disagreement estimate) can be thought of as the norms $\|x_{t}^{i}-x_{t}^{j}\|$, the differences between the iterates $x_{t}^{i}$ and $x_{t}^{j}$. Alternatively, the consensus can be measured with respect to a reference sequence, i.e., ${y_{t}} = \frac{1}{N}\sum^N_{i=1}x^i_{t}$. In particular, we discuss $\|x_{t}^{i}-y_{t}\|$ for any time $t$ as the metrics with respect to the delay $\tau$.
\vspace{5pt}
\begin{lem}(\textbf{\textit{Consensus}})
Let Assumptions 2 and 3 hold. Assume that the delay compensated gradients are uniformly bounded, i.e., there exists a scalar $B>0$, such that 
\[
    \|\mathbf{g}^{dc,r}(\mathbf{x}_{t})\|\leq B,\,\,\ \forall t\geq 0 \,\ and \,\ 0\leq r\leq\tau-1,
\]
Then for all $i\in V$ and $t\geq0$, $\exists \eta > 0$, we have 
\begin{equation}
\mathbb{E}[\|x_{t}^{i}-y_{t}\|]\leq 
\eta\frac{ G+(\tau-1)B\theta_m}{1-\delta_2},
\end{equation}
where $\theta_m=\text{max}\{\theta_{s+1}\}^{t+\tau-1}_{s=t}$, $\delta_2=\text{max}\{\theta_se_2+(1-\theta_s)\tilde{e}_2\}^{t+\tau-1}_{s=0}<1$, where $e_{2}:=e_{2}(W) < 1$ and $\tilde{e}_{2}:=e_{2}(\tilde{W}) < 1$.
\end{lem}

The detailed proof is shown in the Appendix.
Lemma 1 states the consensus bound among agents, which is proportional to the step size $\eta$ and inversely proportional to the gap between the largest and the second-largest magnitude eigenvalues of the equivalent graph $\mathcal{W}$.

\vspace{5pt}
\begin{rem}
One implication that can be made from Lemma 1 is when $\tau = 1$, the consensus bound becomes the smallest, which can be obtained as
$
\frac{\eta G}{1-\delta_2}
$. This bound is the same as obtained already by most decentralized learning (or optimization) algorithms. 
This accordingly implies that the delay compensated gradient or predicting gradient does not 
necessarily require many time steps otherwise more compounding error could be included. Alternatively, $\theta_m = 0$ can also result in such a bound, suggesting that the clipping step dominates in the update. On the other hand, once $\tau\gg 1$ and $\theta_m\neq 0$, the consensus bound becomes worse, which will be validated by the empirical results.
Additionally, if the network is sparse, which suggests $e_{2}\rightarrow 1$ and $\tilde{e}_2\rightarrow 1$, the consensus among agents may not be achieved well and correspondingly the optimality would be negatively affected.
\end{rem}
Most previous works have typically explored the convergence rate on the strongly convex objectives. However, the assumption of strong convexity can be a quite strong condition in most models such that the results obtained may be theoretically instructive and useful. Hence, we introduce a condition that is able to relax the strong convexity but still maintain the similar theoretical property, i.e., Polyak-\L{}ojasiewicz (PL) condition~\cite{karimi2016linear}. The condition is expressed as follows:
A differentiable function $F$ satisfies the PL condition such that there exists a constant $\mu > 0$
\begin{equation}
    \frac{1}{2}\|\nabla F(\mathbf{x})\|^2\geq \mu(F(\mathbf{x})-F^*).
\end{equation}
When $F(\mathbf{x})$ is strongly convex, it also implies the PL condition. However, it is not vice versa. Hence we can arrive at the following results.
\vspace{5pt}
\begin{thm}\label{PL_theorem}
Let Assumptions 1,2 and 3 hold. Assume that the delay compensated gradients are uniformly bounded, i.e., there exits a scalar $B>0$ such that
\begin{equation}
    \|\mathbf{g}^{dc,r}(\mathbf{x}_{t})\|\leq B,\,\,\ \forall t\geq 0 \,\ and \,\ 0\leq r\leq\tau-1,
\end{equation}
and that $\nabla F(\mathbf{x}_t)$ is $\xi_m$-smooth for all $t\geq 0$. Then for the iterates generated by PC-ASGD, when $0<\eta\leq\frac{1}{2\mu\tau}$ and the objective satisfies the PL condition, they satisfy
\begin{equation}
\mathbb{E}[F(\mathbf{x}_{t})-F^{*}]\leq (1-2\mu\eta\tau)^{t-1}(F(\mathbf{x}_1)-F^{*}-\frac{Q}{2\mu\eta\tau})+\frac{Q}{2\mu\eta\tau},
\end{equation}
\begin{equation}
{
\begin{aligned}
    Q &= 2(1-2\mu\eta\tau)G\eta C_1+\frac{\eta^3\xi_mG}{2}\sum_{r=1}^{\tau-1}C_r
+2\eta^2G\gamma_m C_1+G\eta\tau\sigma\\&+\eta^2G(\gamma_m+\epsilon_D+\epsilon+(1-\lambda)G^2)\sum_{r=1}^{\tau-1}C_r
+\eta G^2+\eta^2\gamma_mG\tau C_2\\
\end{aligned}
}
\end{equation}
and
$
C_1=\frac{G+(\tau-1)B\theta_m}{1-\delta_2},
C_r=\frac{2G+(r-1)B\theta_m}{1-\delta_2},
C_2=\frac{2G+(\tau-1)B\theta_m}{1-\delta_2}
$,
$\epsilon_D > 0$ and $\epsilon > 0$ are upper bounds for the approximation errors of the Hessian matrix that are obtained in the Appendix.
\end{thm}
The proof for this theorem is fairly non-trivial and technical. We refer readers to the Appendix for more detail. To simplify the proof, this main result will be divided into several lemmas. One implication from Theorem~\ref{PL_theorem} is that PC-ASGD enables the iterates $\{\mathbf{x}_t\}$ to converge to the neighborhood of $\mathbf{x}^*$, which is $\frac{Q}{2\eta\mu\tau}$. In addition, Theorem~\ref{PL_theorem} shows that the error bound is significantly attributed to network errors caused by the disagreement among agents with respect to the delay and the variance of stochastic gradients. Another implication can be made from Theorem~\ref{PL_theorem} is that the convergence rate is closely related to the delay and the step size such that when the delay is large it may reduce the coefficient, $1-2\mu\eta\tau$, to speed up the convergence. However, correspondingly the upper bound of the step size is also reduced. Hence, there is a tradeoff between the step size and the delay in PC-ASGD.
Theorem~\ref{PL_theorem} also suggests that when the objective function only satisfies the PL condition and is smooth, the convergence to the neighborhood of $\mathbf{x}^*$ in a linear rate can still be achieved. The PL condition may not necessarily imply convexity and hence the conclusion can even apply to some nonconvex functions.

We next investigate the convergence for the non-convex objectives. For PC-ASGD, we show that it converges to a first-order stationary point in a sublinear rate. It should be noted that such a result may not absolutely guarantee a feasible minimizer due to lack of some necessary second-order information. However, for most nonconvex optimization problem, this is generic, though some existing works have discussed about the second-order stationary points~\cite{carmon2018accelerated}, which is out of our investigation scope.
\vspace{5pt}
\begin{thm}\label{Non_convex}
Let Assumptions 1, 2 and 3 hold. Assume that the delay compensated gradients are uniformly bounded, i.e., there exists a a scalar $B>0$ such that for all $T\geq 1$
\begin{equation}
    \|\mathbf{g}^{dc,r}(\mathbf{x}_{t})\|\leq B,\,\,\ \forall t\geq 0 \,\ and \,\ 0\leq r\leq\tau-1,
\end{equation}
and there exists $M$,
\begin{equation}
    \mathbb{E}[\|\mathbf{g}^{dc,r}(\mathbf{x}_t)\|^2]\leq M.
\end{equation}
Then for the iterations generated by PC-ASGD, there exists $0<\eta<\frac{1}{\gamma_m}$, such that 
\begin{equation}
\begin{footnotesize}
\begin{aligned}
\frac{1}{T}\sum_{t=1}^{T}\mathbb{E}[\|\nabla F(\mathbf{x}_t)\|^2]\leq \frac{2(F(\mathbf{x}_1)-F^{*})}{T\eta}+\frac{R}{\eta},\\
\end{aligned}
\end{footnotesize}
\end{equation}
where, $
R=2GC_1+\frac{\tau^2\eta^2\gamma_mM}{2}+\frac{\eta\sigma^2}{2}+\eta\sigma\tau B+2\eta\gamma_m(\tau B+G)C_1,
C_1=\frac{G+(\tau-1)B\theta_m}{1-\delta_2}$.
\end{thm}
\vspace{5pt}
\begin{rem}
Theorem~\ref{Non_convex} states that with a properly chosen constant step size, PC-ASGD is able to converge the iterates $\{\mathbf{x}_T\}$ to the neighborhood of a stationary point $\mathbf{x}^*$ in a rate of $O(T^{-1})$, whose radius is determined by $\frac{R}{\eta}$. Additionally, based on $R$, we can know that the error bound is mainly caused by the variance of stochastic gradients and the network errors. One can also observe that the error bound increases when the delay becomes larger. As the length of delay can have an impact on the prediction steps used in the delay compensated gradient, a short term prediction may help alleviate the negative effect caused by the delay. Otherwise, the compounding error in the delay compensated gradient could deteriorate the performance of the algorithm.
\end{rem}

\section{Experiments} 
\label{secExperiments}
\subsection{Practical Variant}
So far, we have analyzed theoretically in detail how the proposed PC-ASGD converges with some mild assumptions. 
In practical implementation, we need to choose a suitable $\theta_t$ to enable the training fast with clipping steps and allow the unreliable neighbors to be involved in training with predicting steps. In this context, we develop a heuristic practical variant with a criterion for determining the tradeoff parameter value. Intuitively, if the delay messages from the unreliable neighbors do not influence the training negatively, they should be included in the prediction. This can be determined by the comparison with the algorithm without making use of these messages. 
The criterion is shown as follows:

\begin{equation}\label{criteria}
x_{i}^{t+1}= 
\left\{
\begin{array}{rcl} 
x_{t+1,pre}^{i}& {\frac{\langle x_{t+1,pre}^{i}-x^i_t,g_i(x^i_t)\rangle}{\|x_{t+1,pre}^{i}-x^i_t\|}}\geq&{\frac{\langle x_{t+1,cli}^{i}-x^i_t,g_i(x^i_t)\rangle}{\|x_{t+1,cli}^{i}-x^i_t\|}}\\
x_{t+1,cli}^{i} & o.w.\\
\end{array}\right.
\end{equation}

where we choose the \textit{cosine distance} to compare the distances for predicting and clipping steps. The prediction step is selected if it has the larger cosine distance, which implies that the update due to the predicting step yields the larger loss descent. Otherwise, the clipping step should be chosen by only trusting reliable neighbors.  
Our practical variant with this criterion still converges since we just set $\theta_t$ as $0$ or $1$ for each iteration and the previous analysis in our paper still holds. To facilitate the understanding of predicting and clipping steps, in the following experiments, we also have two other variants P-ASGD and C-ASGD. While the former corresponds to an ``optimistic" scenario to only rely on the predicting step, the latter presents a ``pessimistic" scenario by dropping all outdated agents. Both of variants follow the same convergence rates induced by PC-ASGD.
The specific algorithm is showed as Algorithm \ref{pcasgd_1}.

\renewcommand{\algorithmicrequire}{\textbf{Input:}}
 \renewcommand{\algorithmicensure}{\textbf{Output:}}
 \begin{algorithm}[htb]  
 \caption{PC-ASGD-PV}
 \label{pcasgd_1}
 \begin{algorithmic}[1] 
 \REQUIRE number of agents \(N\), learning rate $\eta > 0$,  agent interaction matrices \(\underline{W}\), \(\underline{\tilde{W}}\), number of epochs $T$
 \ENSURE the models' parameters in agents $x_{T}^i$,$i=1,2,...N$
 \STATE \textbf{Initialize} all the agents' parameters $x_{0}^i$, $i=1,2,...N$
 \STATE Do broadcast to identify the clusters of reliable agents and the delay $\tau$
 \STATE $t=0$
 \WHILE{$epoch\;t < T$} 
 \FOR{each agent $i$}
 \STATE Predicting Step: $x^i_{t+1, pre} = \sum_{j\in \mathcal{R}}w_{ij}x^j_t - \eta g_i(x^i_t)+\sum_{k\in \mathcal{R}^c}w_{ik}(x^k_{t-\tau} - \eta g^{dc}_k(x^k_{t-\tau}))$
 \STATE Clipping Step: $x^i_{t+1, cli} = \sum_{j\in \mathcal{R}}\tilde{w}_{ij}x^j_t - \eta g_i(x^i_t)$
 \STATE $\Delta_{pre}=x^i_{t+1,pre}-x^i_t;\;\Delta_{cli}=x^i_{t+1,cli}-x^i_t$
 \IF {$\frac{\langle \Delta_{pre}, g_i(x^i_t)\rangle}{\|\Delta_{pre}\|} \geq \frac{\langle \Delta_{cli}, g_i(x^i_t)\rangle}{\|\Delta_{cli}\|}$}
 \STATE $x^i_{t+1} = x^i_{t+1,pre}$
 \ELSE
 \STATE $x^i_{t+1} = x^i_{t+1,cli}$
 \ENDIF
 \ENDFOR
 \STATE $t=t+1$
 \ENDWHILE
 \end{algorithmic}
 \end{algorithm}

\subsection{Distributed Network and Learning Setting}

\textbf{Models and Data sets}.  Decentralized asynchronous SGD (D-ASGD) is adopted as the baseline algorithm. 
Two deep learning structures, PreResNet110 and DenseNet (noted as \emph{model 1} and \emph{model 2}), are employed. The detailed model structures are illustrated in the Appendix. CIFAR-10 and CIFAR-100 are used in the experiments following the settings in \cite{Krizhevsky2009LearningML}.
The training data is randomly assigned to each agent, and the parameters of the deep learning structure are maintained within each agent and communicated with the predefined delays. The testing set is utilized for each agent to verify the performance, where our metric is the average accuracy among the agents. 6 runs are carried out for each case and the mean and variance are obtained and listed in Table \ref{performance comparison}. 

\textbf{Delay setting}. The delay is set as $\tau$ as discussed before, which means the parameters received from the agents outside of the reliable cluster are the ones that were obtained $\tau$ iterations before. For \emph{model 1} and \emph{model 2}, $\tau$ is both fixed at 20 to test the performances of different algorithms including our different variants (D-ASGD, P-ASGD, C-ASGD, and PC-ASGD) and baseline algorithms in Section \ref{subsecPerformance} and \ref{subsecnetworksize}. We also try to exploit its impact in Section \ref{subimpactdelay}.  

\textbf{Distributed network setting}. A distributed network (noted as \emph{distributed network 1}) with $8$ agents (nodes) in a fully connected graph is first applied with \emph{model 1} and \emph{model 2},  and 2 clusters of reliable agents are defined within the graph consisting of 3 agents and 5 agents, respectively. Then two distributed networks (with 5-agent and 20-agent, respectively) are used for scalability analysis, noted as \emph{distributed network 2} and \emph{distributed network 3}, respectively. For \emph{distributed network 2}, we construct 2 clusters of reliable agents with 3 and 2 agents. In \emph{distributed network 3}, four clusters are formed and 3 clusters consist of $6$ agents while the rest has $2$ agents.

\begin{figure*}[htbp] 
    \centering 
    \subfigtopskip=2pt 
    \subfigbottomskip=2pt 
    \subfigcapskip=-5pt
    \subfigure[\scriptsize{DenseNet CIFAR-10}]{
        \label{level.sub.2.1}        \includegraphics[width=0.35\linewidth]{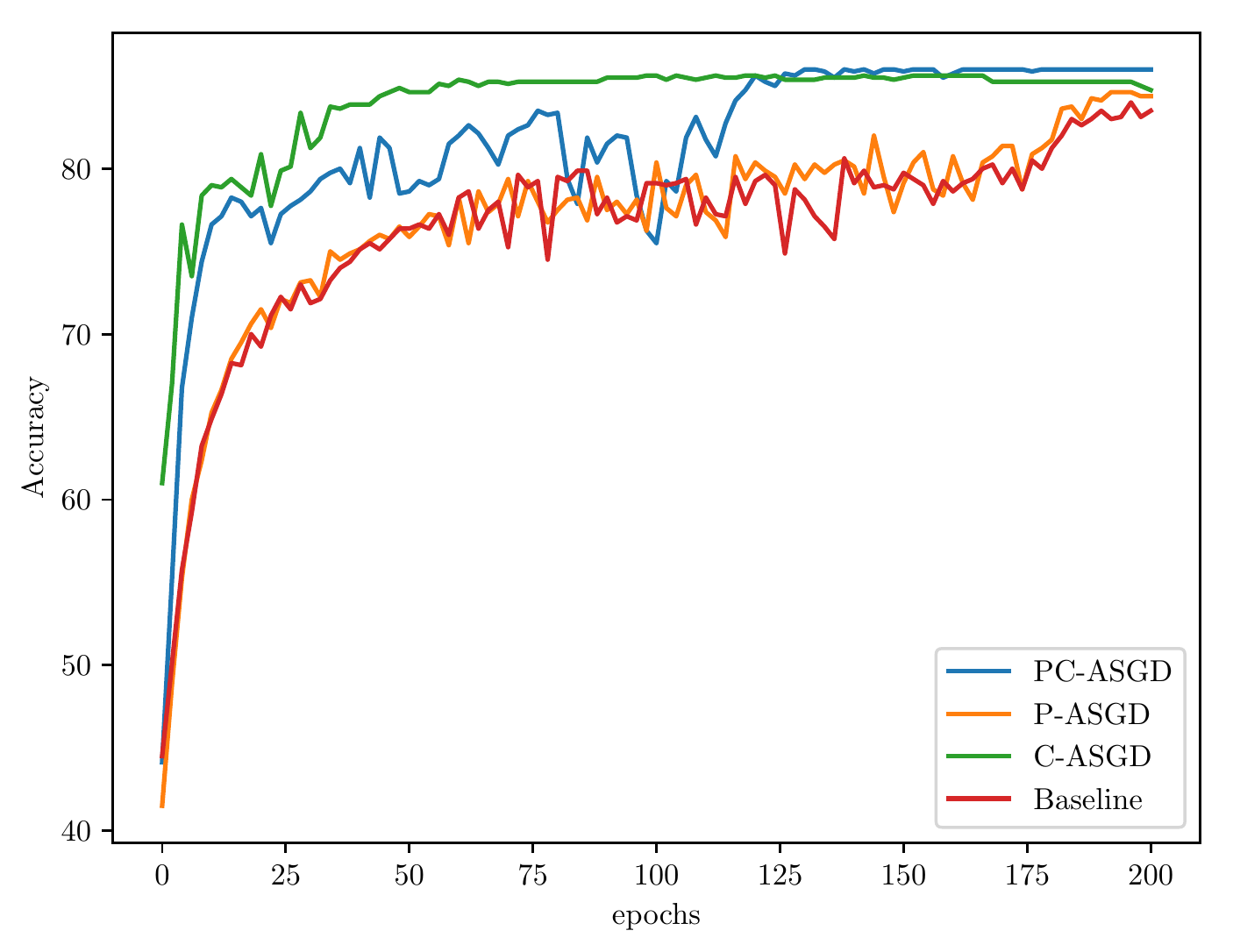}}
    \subfigure[\scriptsize{DenseNet CIFAR-100}]{
        \label{level.sub.2.2}
        \includegraphics[width=0.35\linewidth]{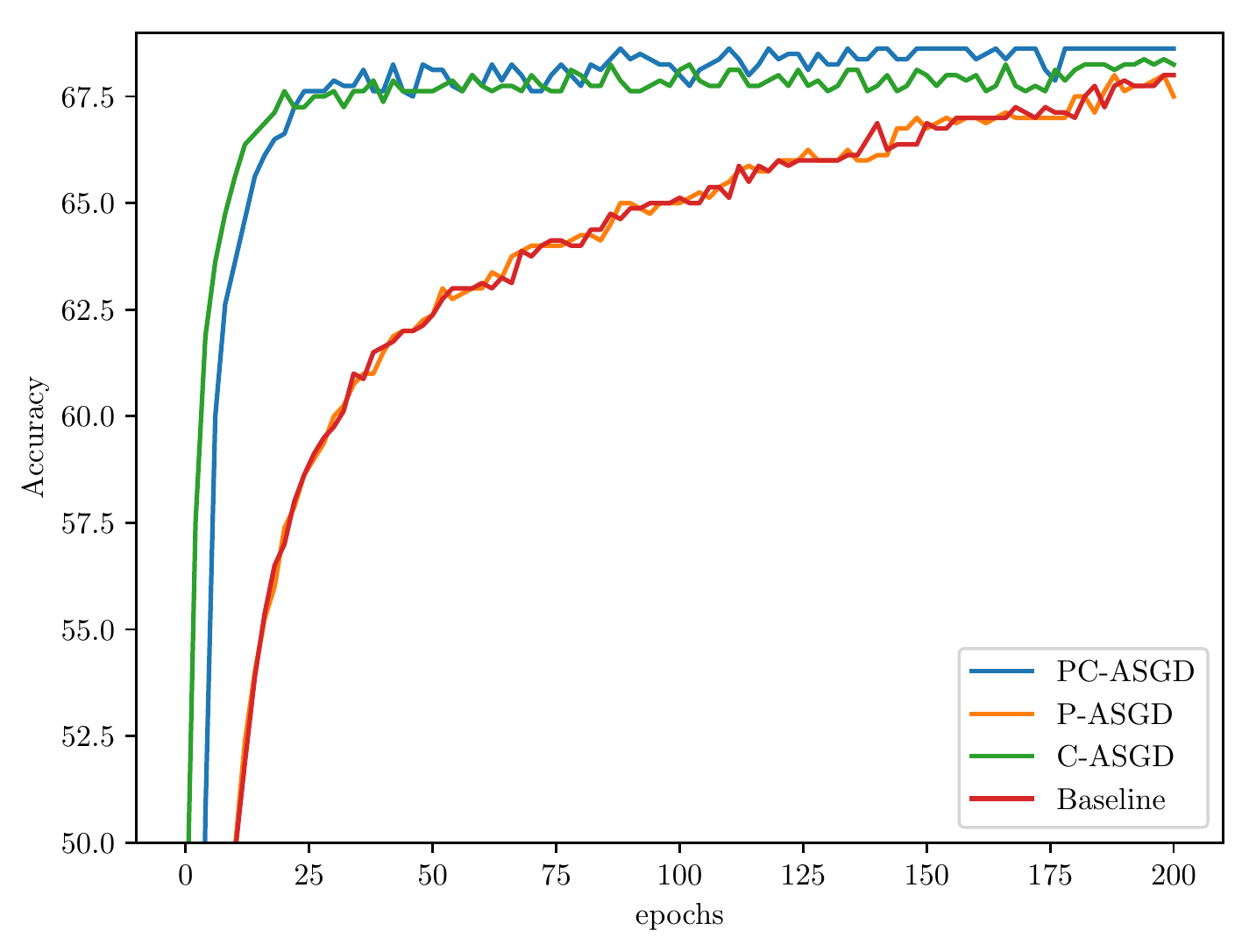}}
    \subfigure[\scriptsize{PreResNet110 CIFAR-10}]{
        \label{level.sub.2.3}
        \includegraphics[width=0.35\linewidth]{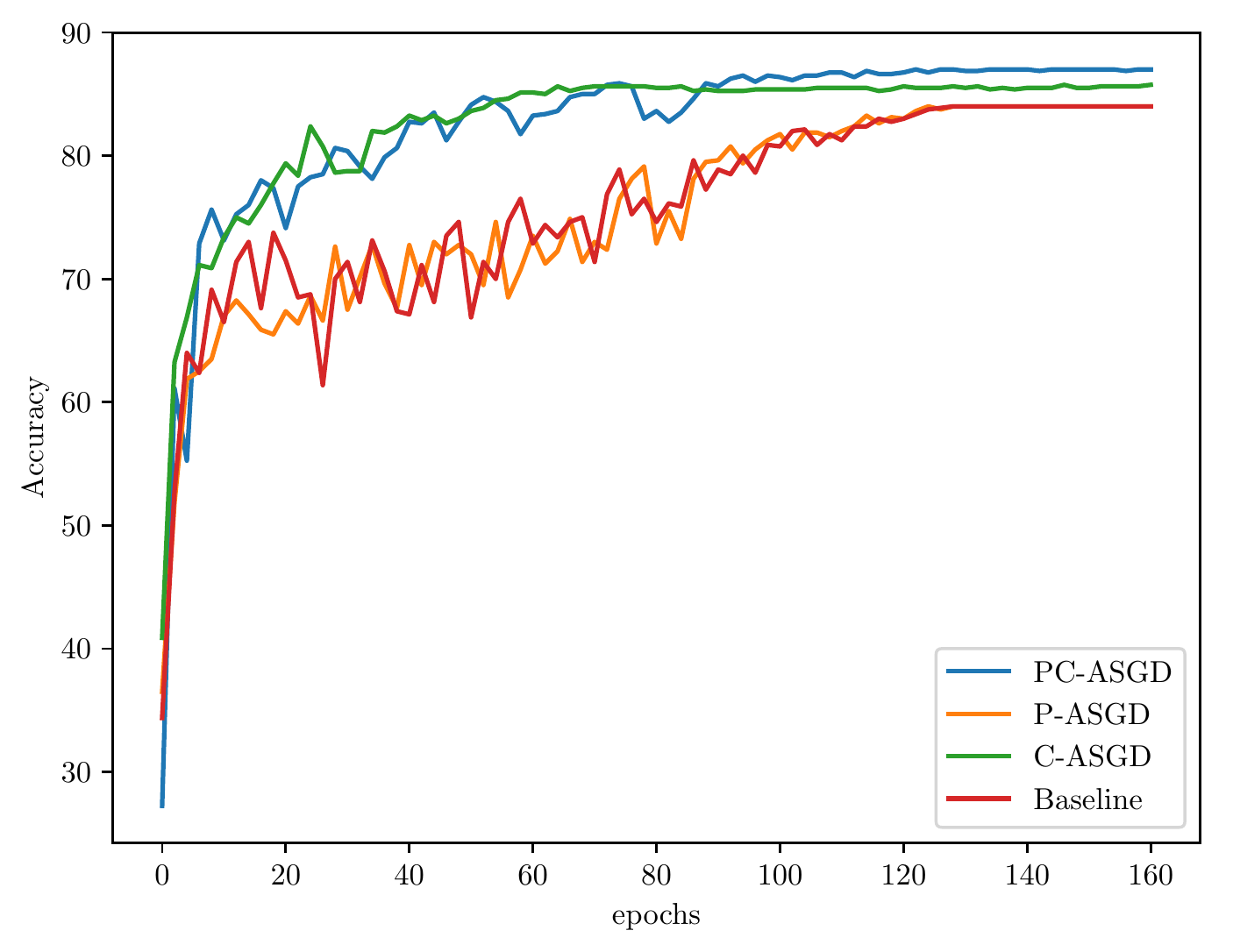}}
    \subfigure[\scriptsize{PreResNet110 CIFAR-100}]{
        \label{level.sub.2.4}
        \includegraphics[width=0.35\linewidth]{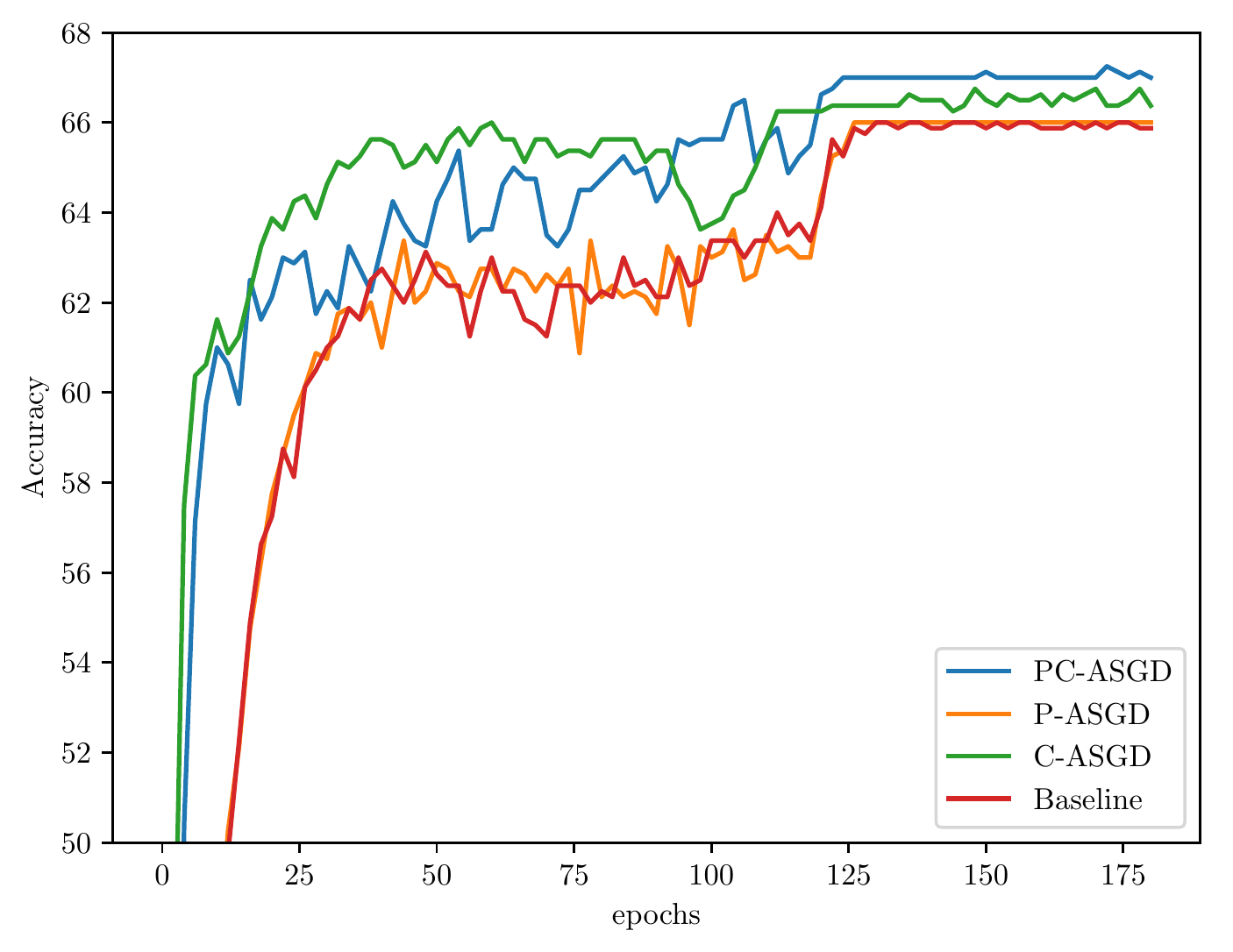}}
    \caption{Testing accuracy on CIFAR-10 and CIFAR-100.}
    \label{level}
\end{figure*}
\vspace{5pt}

\subsection{Performance Evaluation}
\label{subsecPerformance}

The testing accuracies on the CIFAR-10 and CIFAR-100 data sets with \emph{model 1} and \emph{model 2} in \emph{distributed network 1} are shown in Fig. \ref{level}. It shows that the proposed PC-ASGD outperforms the other single variants and it presents an accuracy increment greater than $2.3\%$ (nearly $4\%$ for DenseNet with CIFAR-10) compared to the baseline algorithm. For other variants P-ASGD or C-ASGD, the testing accuracies are also higher than that of the baseline algorithm. Moreover, PC-ASGD shows faster convergence than P-ASGD as the updating rule overcomes the staleness, and achieves better accuracy than the C-ASGD as it includes the messages from the unreliable neighbors. This is consistent with the analysis in this work. We also show the detailed results of both \emph{distributed network 1} and \emph{distributed network 3} in Table \ref{tab20}.

\begin{table*}[htbp]
\caption{Performance evaluation of PC-ASGD on CIFAR-10 and CIFAR-100}
\centering
\begin{footnotesize}
\label{tabResu}
\setlength{\tabcolsep}{1.2mm}{	
\begin{tabular}{c c c c c c c c} 
\hline
\multicolumn{8}{c}{8 agents}\\ \hline
        \multirow{2}[2]{*}{Model \& dataset}   & \multicolumn{2}{c}{PC-ASGD}  & \multicolumn{2}{c}{P-ASGD} & \multicolumn{2}{c}{C-ASGD}  & Baseline\\ 
                      & acc. \scriptsize{(\%)} & o.p. \scriptsize{(\%)} & acc. (\%) & o.p. \scriptsize{(\%)} & acc.\scriptsize{(\%)} & o.p. \scriptsize{(\%)} & acc. \scriptsize{(\%)} \\ \hline
Pre110, CIFAR-10 & $\mathbf{87.3\pm1.1}$  & $\mathbf{3.3\pm1.1}$ & $84.9\pm0.9$   & $0.9\pm0.9$ & $86.0\pm1.0$   & $2.0\pm1.0$ & $84.0\pm0.3$  \\ \hline
Pre110, CIFAR-100 & $\mathbf{67.4\pm1.4}$  & $\mathbf{3.1\pm1.9}$ & $64.8\pm 1.3$  & $1.3\pm1.5$  & $66.4\pm1.2$   & $1.9\pm1.6$ & $64.5\pm1.5$  \\ \hline
Des, CIFAR-10  & $\mathbf{86.9\pm0.9}$    & $\mathbf{3.6\pm1.8}$ & $84.4\pm0.6$    & $1.0\pm1.5$  & $85.9\pm0.9$   & $2.7\pm1.7$ & $83.3\pm0.9$  \\ \hline
Des, CIFAR-100  & $\mathbf{68.6\pm0.6}$   & $\mathbf{2.3\pm1.7}$ & $66.8\pm1.5$   & $1.6\pm1.6$  & $66.8\pm1.6$   & $1.8\pm1.6$ & $66.1\pm1.9$  \\ \hline
\hline
\multicolumn{8}{c}{20 agents}\\ \hline

        \multirow{2}[2]{*}{Model \& dataset}   & \multicolumn{2}{c}{PC-ASGD}  & \multicolumn{2}{c}{P-ASGD} & \multicolumn{2}{c}{C-ASGD}  & Baseline\\ 
                      & acc. \scriptsize{(\%)} & o.p. \scriptsize{(\%)} & acc. (\%) & o.p. \scriptsize{(\%)} & acc.\scriptsize{(\%)} & o.p. \scriptsize{(\%)} & acc. \scriptsize{(\%)} \\ \hline
Pre110, CIFAR-10 & $\mathbf{84.7\pm0.9}$  & $\mathbf{4.2\pm1.0}$ & $83.3\pm 0.9$   & $2.7\pm0.9$ &$82.5\pm 1.0$    & $1.9\pm1.4$ & $80.4\pm 0.7$ \\ \hline
Pre110, CIFAR-100 & $\mathbf{62.4\pm0.8}$  & $\mathbf{3.3\pm2.0}$ & $61.7\pm 1.0$  & $2.0\pm1.6$  & $61.5\pm1.0$   & $2.5\pm2.3$ & $59.3\pm1.7$  \\ \hline
Des, CIFAR-10  & $\mathbf{82.9\pm0.9}$     & $\mathbf{2.4\pm0.9}$ & $82.0\pm0.7$    & $1.4\pm1.3$  & $81.8\pm0.6$   & $1.8\pm1.0$ & $80.1\pm0.9$  \\ \hline
Des, CIFAR-100  & $\mathbf{64.5\pm0.7}$   & $\mathbf{3.8\pm1.7}$& $62.5\pm1.3$   & $2.9\pm2.0$  & $62.0\pm1.5$   & $1.3\pm1.4$ & $60.4\pm1.7$  \\ \hline
\end{tabular}\label{tab20}}
\end{footnotesize}
\\
\begin{tablenotes}
\item \scriptsize{acc.--accuracy, o.p.--outperformed comparing to baseline}. 
\end{tablenotes}
\end{table*}

We then compare our proposed algorithm with other delay-tolerant algorithms, including the baseline algorithm D-ASGD (aka AD-PSGD), DC-s3gd \cite{rigazzi2019dc-s3gd:}, DASGD with IS  \cite{du2020asynchronous}, and Adaptive Braking \cite{venigalla2020adaptive}.
The \emph{distributed network 1} is applied for the comparisons. 
\begin{table*}[tbp]
\caption{Performance comparison for different delay tolerant algorithms}
\centering
\setlength{\tabcolsep}{1.2mm}{	
\begin{tabular}{l|l|l|l|l}
\hline
Model \& dataset & Pre110,CIFAR-10 & Pre110,CIFAR-100 & Des,CIFAR-10 & Des,CIFAR-100 \\ \hline
PC-ASGD        &        $\mathbf{87.3 \pm 1.1} $         &     $\mathbf{67.4\pm 1.4}$             &    $\mathbf{86.9 \pm 0.6}$          &    $\mathbf{68.6\pm 0.6}$           \\ \hline
AD-PSGD \cite{lian2017asynchronous}         & $84.0 \pm 0.3$         &     $64.5\pm 1.5$             &    $83.3 \pm 0.9$          &    $66.1\pm 1.9$          \\ \hline
DC-s3gd \cite{rigazzi2019dc-s3gd:}          & $86.3 \pm 0.8$         &     $63.5\pm 1.7$             &    $85.7 \pm 0.8$          &    $66.2\pm 1.3$      \\ \hline
DASGD with IS  \cite{du2020asynchronous}    & $85.0 \pm 0.3$         &     $64.6\pm 1.2$             &    $84.6 \pm 0.4$          &    $66.2\pm 0.8$      \\ \hline
Adaptive Braking \cite{venigalla2020adaptive}          & $86.8 \pm 0.9$         &     $66.5\pm 1.2$             &    $85.3 \pm 1.0$          &    $67.3\pm 1.1$    \\ \hline
\end{tabular}\label{performance comparison}}
\end{table*}
From the Table~\ref{performance comparison},
the proposed PC-ASGD obtains the best results in the four cases. It should be noted that some of above listed algorithms are not designed specifically for this kind of peer-to-peer applications (e.g., Adaptive Braking) or may not consider the modelling of severe delays in their works (e.g., DASGD with IS and DC-s3gd). In this context, they may not perform well in the test cases. 

\subsection{Impacts of Different Delay Settings}\label{subimpactdelay}

\begin{figure*}[htbp]
    \centering
    \includegraphics[width = 12 cm]{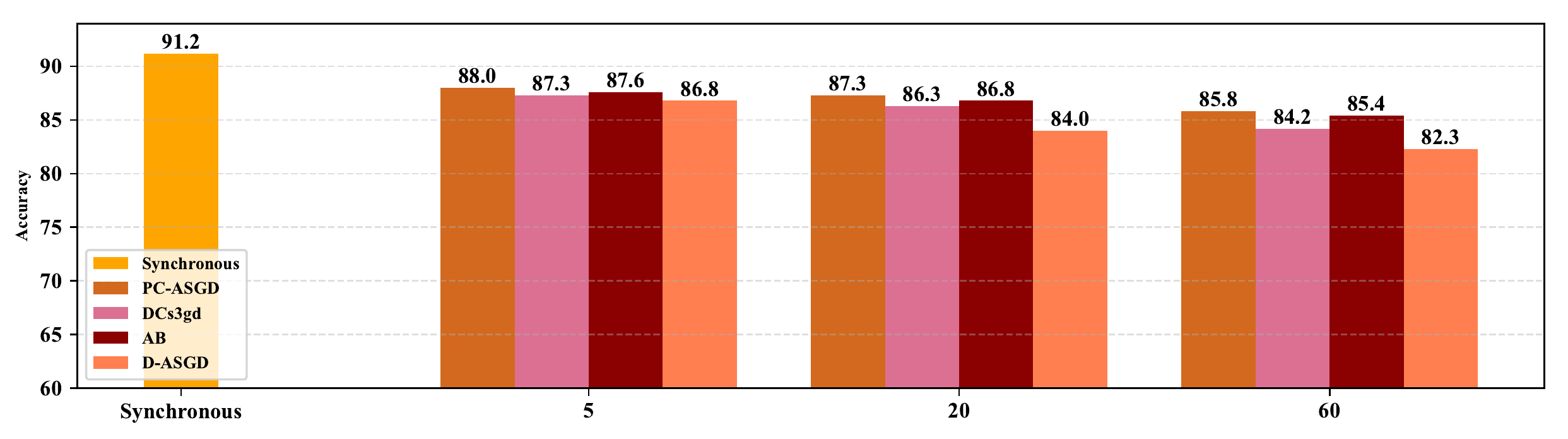}
    \caption{Performance evaluation for different steps of delay.}
    \label{delay}
\end{figure*}

To further show our algorithm's effectiveness, we also implement experiments with different delays.
As discussed above, a more severe delay could cause significant drop on the accuracy. More  
numerical studies with different steps of delay are carried out here. The delays are set as $5,20,60$ with our PreResNet110 (\emph{model 1}) of 8 agents (synchronous network without delay is also tested). We use CIFAR-10 in the studies and the topology is \emph{distributed network 1}. The results are shown in Fig. \ref{delay}. 

\vspace{3pt}
We can find out as the delay increases, the accuracy decreases. 
For the synchronous setting, the testing accuracy is close to that in the centralized scenario
\cite{gitcode} but with higher batch size. When the delay is $60$, the accuracy for the D-ASGD reduces significantly, and this validates that the large delay significantly influences the performance and causes difficulties in the training process. However, the delays are practical in the real implementations such as industrial IoT platforms. For our proposed PC-ASGD, it outperforms other algorithms in all cases with different delays. Moreover, the accuracy drop is relatively smaller in cases with larger delays, which suggests that PC-ASGD is more robust to different communication delays.

\subsection{Impacts of Network Size}\label{subsecnetworksize}

\begin{figure*}
    \centering
    \includegraphics[width = 12 cm]{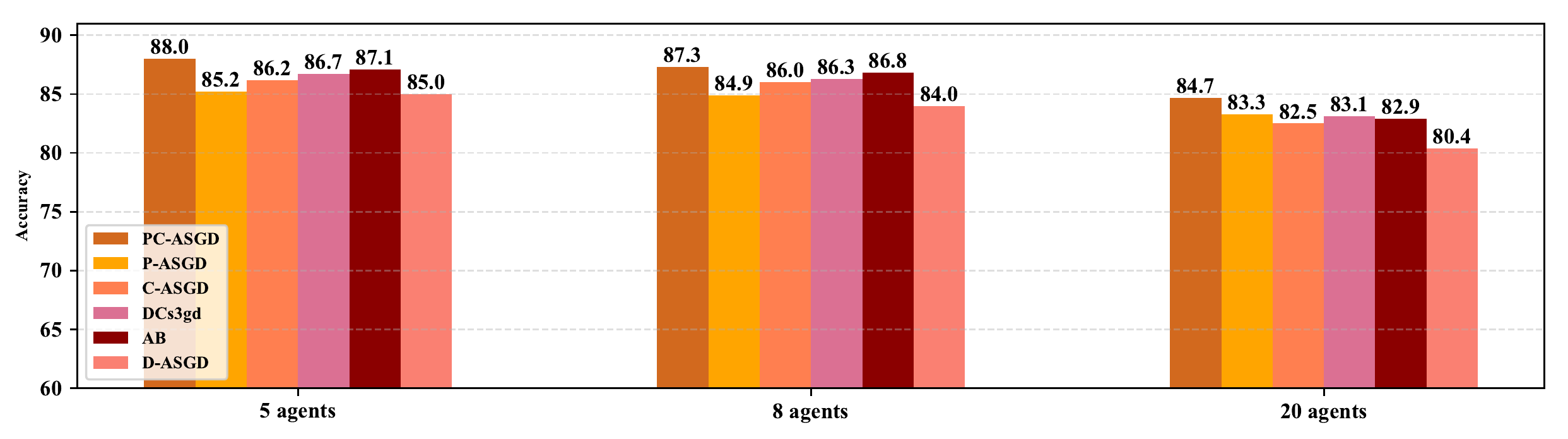}
    \caption{Performance evaluation for different numbers of agents.}
    \label{multi-agent}
\end{figure*}

For evaluating the performances in different structure sizes of distributed networks, \emph{distributed network 2} and \emph{distributed network 3} follow the same setting as in the \emph{distributed network 1} (delay $\tau=20$, \emph{model 1}, CIFAR-10). 
The results are shown in Fig. \ref{multi-agent}.
According to both Table \ref{tab20} and Fig. \ref{multi-agent}, as the number of agents increases, the accuracy decreases. It shows that the large size of the network has negative impact on the training. Our proposed PC-ASGD outperforms all other approaches, which further validates the efficacy and scalability of the proposed algorithm.

\subsection{Numerical Studies on \texorpdfstring{$\theta$}{theta} Assignments}
We also conduct empirical studies about the different choices for $\theta$. As we mentioned above, a practical variant is applied for $\theta$, where we intend to form a strategy to determine if the received information (parameters of the deep learning models) is outdated or not. Here, different assignment rules for $\theta$ are tested and compared. 
\emph{Model 1} is applied, by using CIFAR-10 and the 8 agents system with 3 and 5 agents (\emph{distributed network 1}).

First, $\theta$ is fixed as $0.3,0.5,0.7$ (denoted as \emph{f1, f2, f3}), respectively. Then we determine the $\theta$ as $0,1$ randomly with fixed probability in each round with $0.3,0.5,0.7$ (denoted as \emph{p1, p2, p3}). We also try the fully uniformly random assigned $\theta$ in each round (denoted as \emph{r1}). 
\begin{table}[htbp]
\caption{Mean Performance for Different $\theta$ assignment for Pre110, CIFAR-10}
\centering
\begin{tabular}{l|l|l|l}
\hline
Method\textbackslash{}Parameters & \emph{f1/p1} & \emph{f2/p2} & \emph{f3/p3} \\ \hline
$\theta$ Fixed                   & $86.3$ & $85.0$  & $84.5$       \\ \hline
$\theta$ Bool randomly           & $85.6$ & $85.0$  & $84.1$       \\ \hline
$\theta$ randomly                & \multicolumn{3}{c}{$85.2$} \\ \hline
PC-ASGD-PV                          & \multicolumn{3}{c}{$\mathbf{87.3}$} \\ \hline
D-ASGD(Baseline)                 & \multicolumn{3}{c}{$84.0$} \\ \hline
\end{tabular}\label{theta assignment}
\end{table}
The results are listed in Table \ref{theta assignment}. The PC-ASGD-PV obtains the best performance which implies that the trade-off between the predicting step and the clipping step in the Algorithm 2 is proper and plays an important role in the convergence process.  
With the fixed $\theta$ (first row `$\theta$ fixed'), the experimental results show that the optimal ratio between the predicting step and clipping step is 0.3 in this case. And this suggests that more clipping steps are better.

For the \emph{p1, p2, p3} cases (second row $\theta$ Bool randomly), the experimental results show that the optimal probability between the predicting step and clipping step is 0.3. This is consistent with the fixed $\theta$ case. 
Compared with the fix $\theta$ setting, picking $0, 1$ for the $\theta$ in a predefined probability performs worse. The randomness still help the convergence process but is not as good as the fix $\theta$ setting. For the random $\theta$, the randomness helps the convergence process. However, there exists a optimal $\theta$ for every case and the randomness is not able to get the best performance.
The baseline D-ASGD gets the worst performance, which shows the predicting and clipping steps are helpful for the scenarios with delays in the distributed network. This also provides us the necessity of the additional time cost for the predicting and clipping steps.
It should be noted that optimizing the selection of $\theta$ is beneficial and we can set $\theta$ as binary or non-binary (continuous). The binary setting with the strategy in Algorithm 2 is straightforward and performs well in this work.

\begin{figure}[htbp]
    \centering
    \includegraphics[width = 6
    cm]{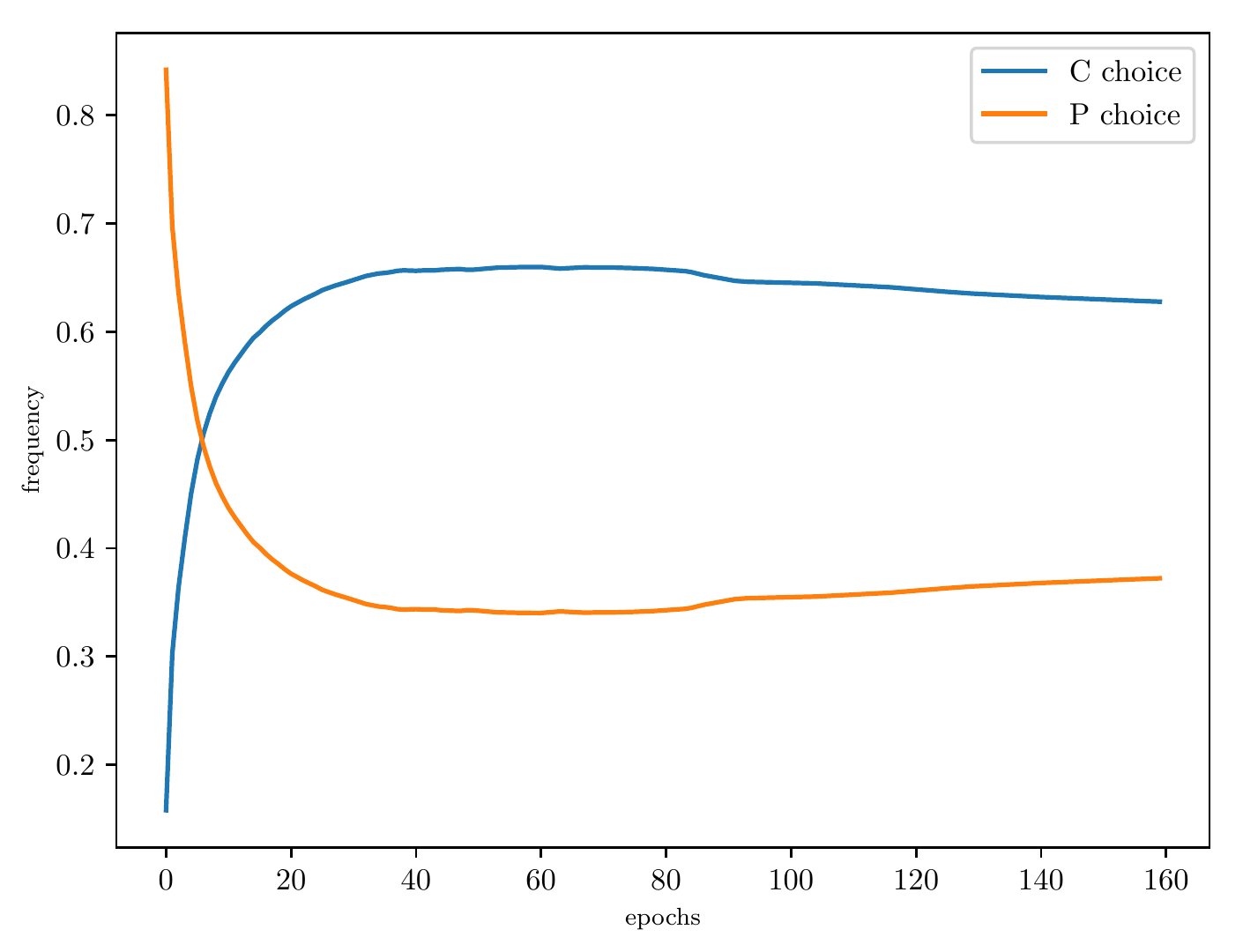}
    \caption{Predicting and clipping steps choices changing with epochs.}
    \label{pcchoice}
\end{figure}

To further explore the connection between the $\theta$ selection and the binary strategy in our algorithm, the occurrence of choosing the predicting step or clipping step in PC-ASGD-PV is collected and shown in  Fig. \ref{pcchoice}. The frequency for the predicting or clipping step choice tends to stabilize with the epochs increasing to the value of $0.625$ and $0.375$. This is consistent with the fixed $\theta$ experiments (where the optimal ratio between the predicting step and clipping step is $0.3$, compared to $0.5$ and $0.7$.)

\subsection{Time Cost Comparison}
\begin{figure}
    \centering
    \includegraphics[width = 6cm]{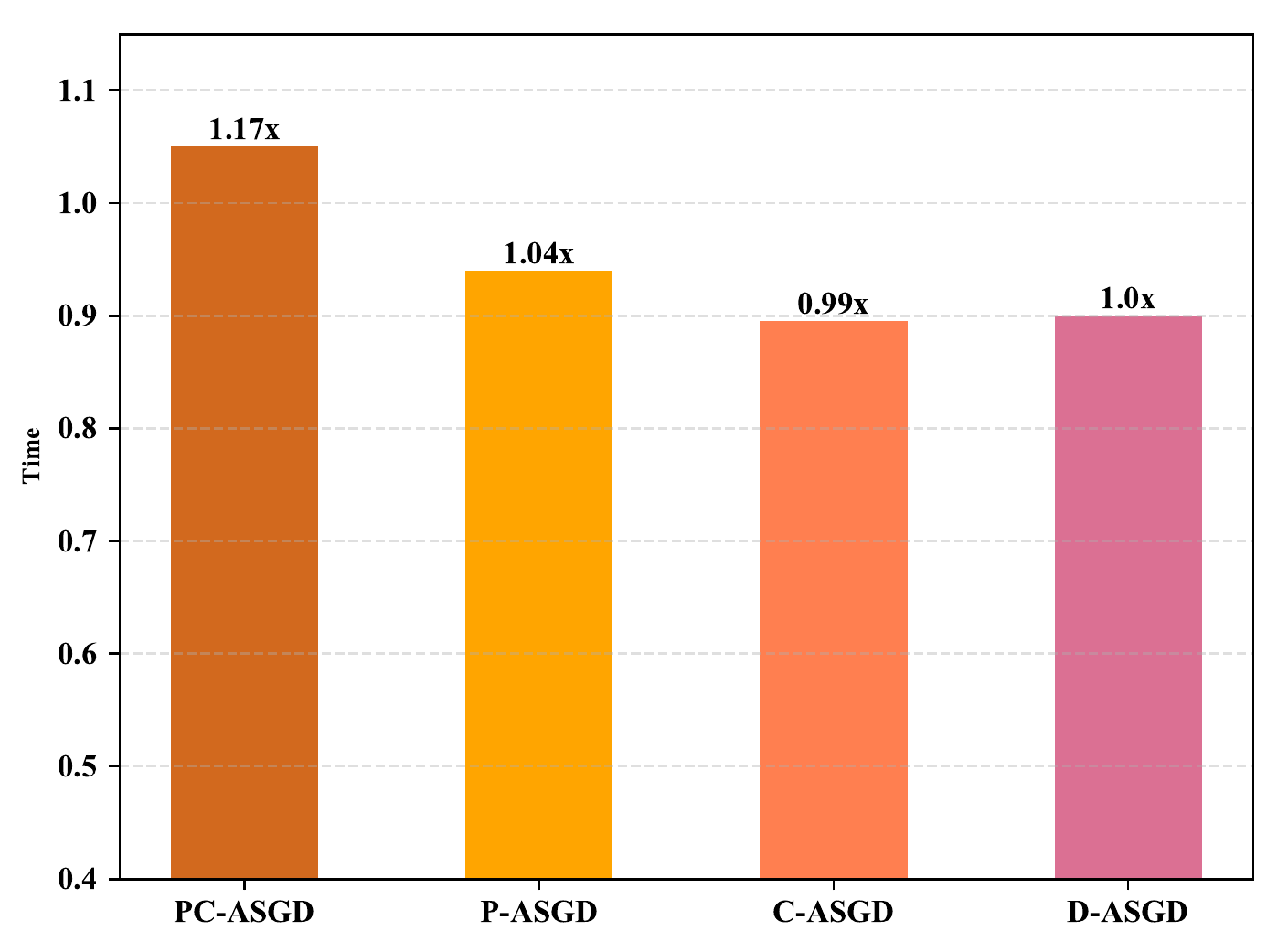}
    \caption{Average time costs for different methods (per epoch).}
    \label{time}
\end{figure}

The time cost for the presented algorithm is compared with the baseline algorithm (D-ASGD), P-ASGD, and C-ASGD. 
The average time costs for \emph{model 1} with CIFAR-10 in \emph{distributed network 1} are collected and shown in
Fig. \ref{time}. 

We observe that the extra time costs for the predicting and clipping steps and additional criterion are not large, although there are still 17\% more costs comparing to D-ASGD. 
Therefore, we need to consider the trade-off before implementing the proposed algorithm. However, with the improvement of the local computing resources and the architecture design, the extra time cost might be acceptable as of the gains in the performance. 
Moreover, the extra time cost is not changed with the delay, while the boosting in the performance is more significant in large delays (as shown in Fig. \ref{delay}). It means that our algorithm could be more applicable in the distributed network with various delays, and this is realistic in industrial IoT systems where the computing resources vary remarkably among the agents and the data in each agent also differs significantly.

\subsection{Verification Using Simple Functions}
Finally, we present two examples to verify our constructed theoretical analysis.
We construct a network involving three agents. We also set two reliable clusters with 1 and 2 agents, respectively. 
We leverage two nonconvex functions, i.e., Rastrigin and Rosenbrock to test the
performance of our proposed framework. Though Rastrigin and Rosenbrock functions are simple nonconvex problems, they have been used widely to test the performance for many numerical optimizers~\cite{mishra2006some}. We randomly sample batch when local training in each agent. We set a fixed step size according our Theorem 2 as 0.008. The number of iterations is set 500 for each case.
\begin{figure*}[htbp]
\centering
\subfigure[Convergence trajectories for Rosenbrock function]{\label{ro}
\includegraphics[width=0.35\linewidth]{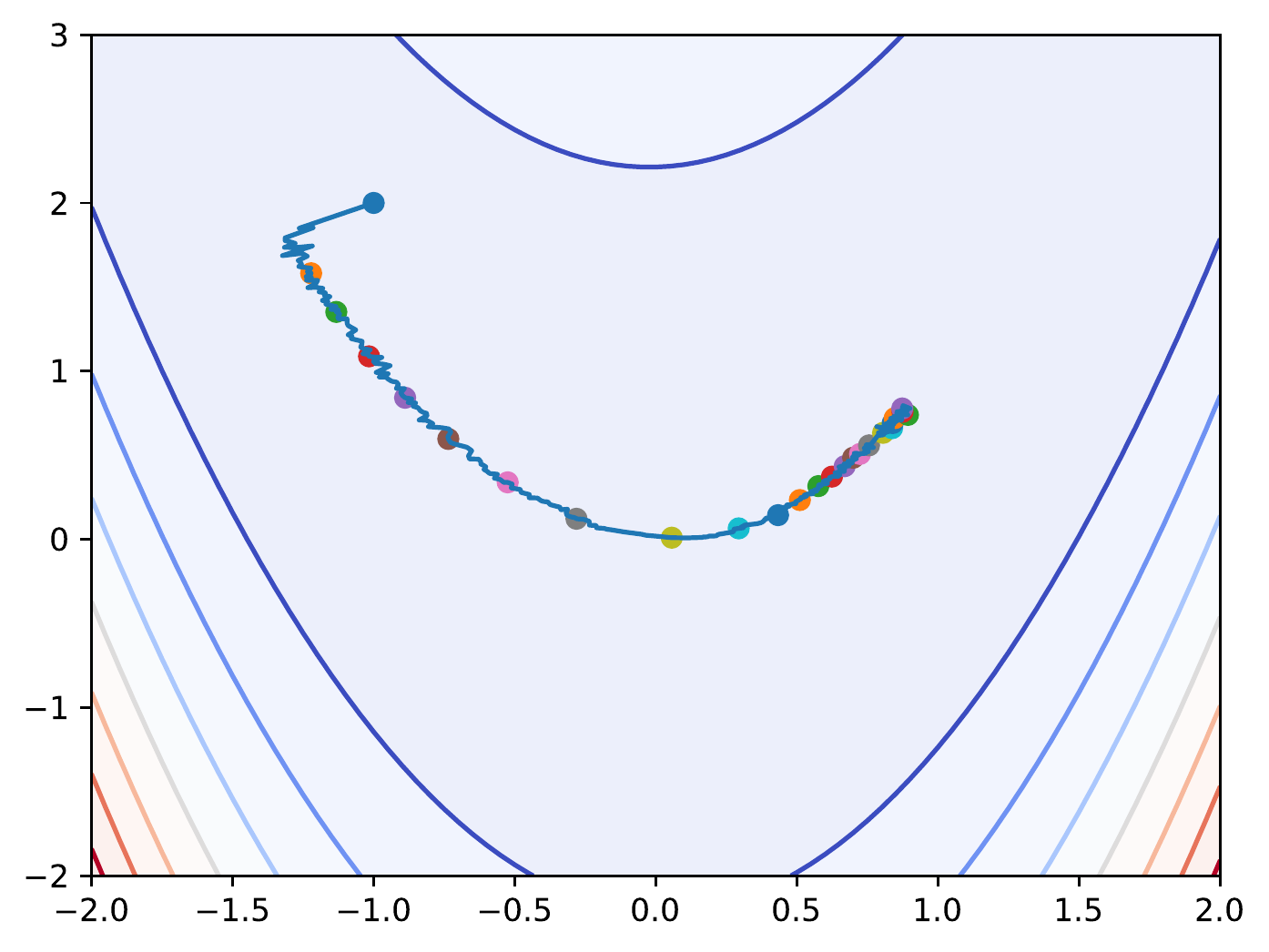}}
\subfigure[Average gradients bound verification in Rosenbrock function]{\label{rob}
\includegraphics[width=0.35\linewidth]{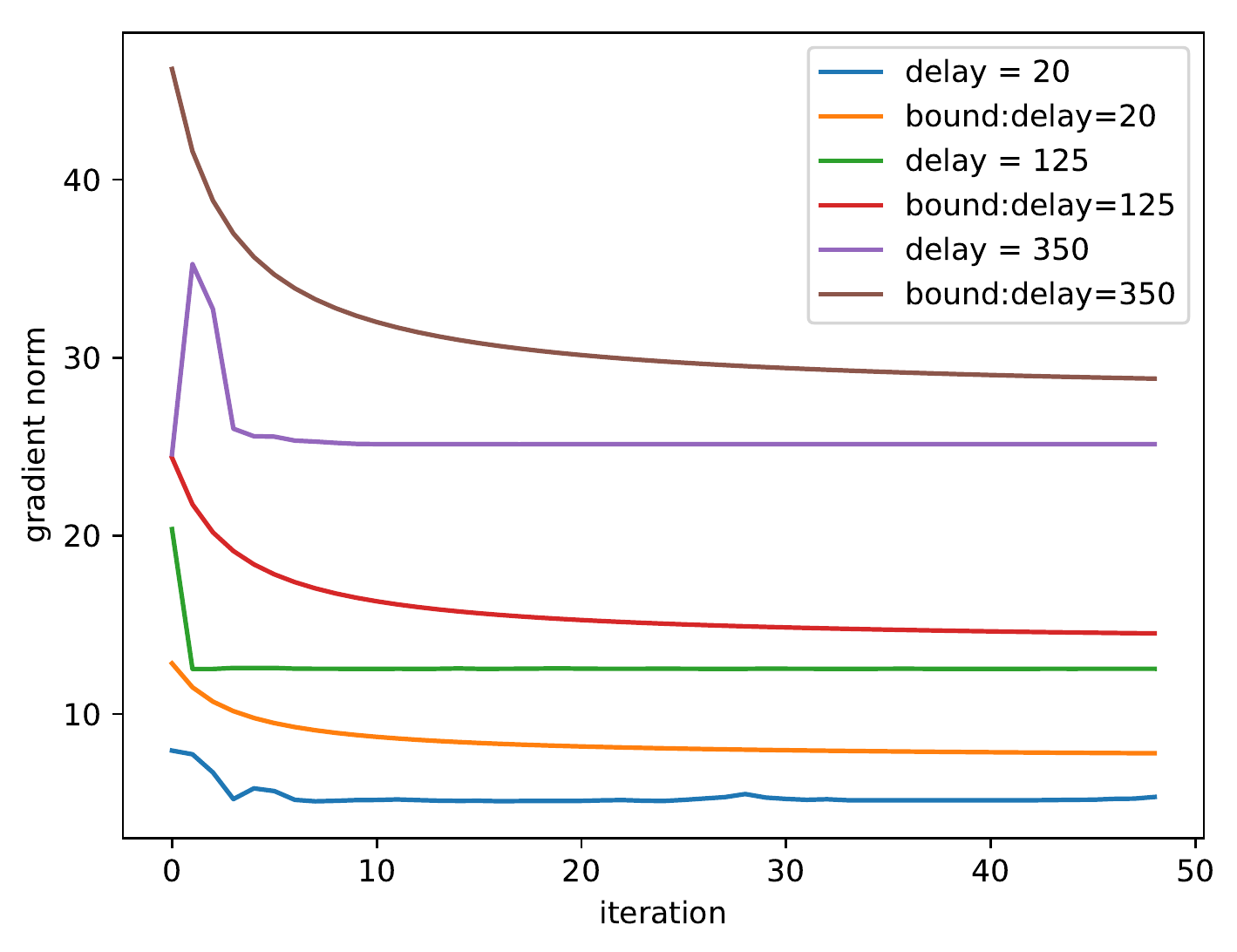}}
\subfigure[Convergence trajectories for Rastrigin function]{\label{rt}
\includegraphics[width=0.35\linewidth]{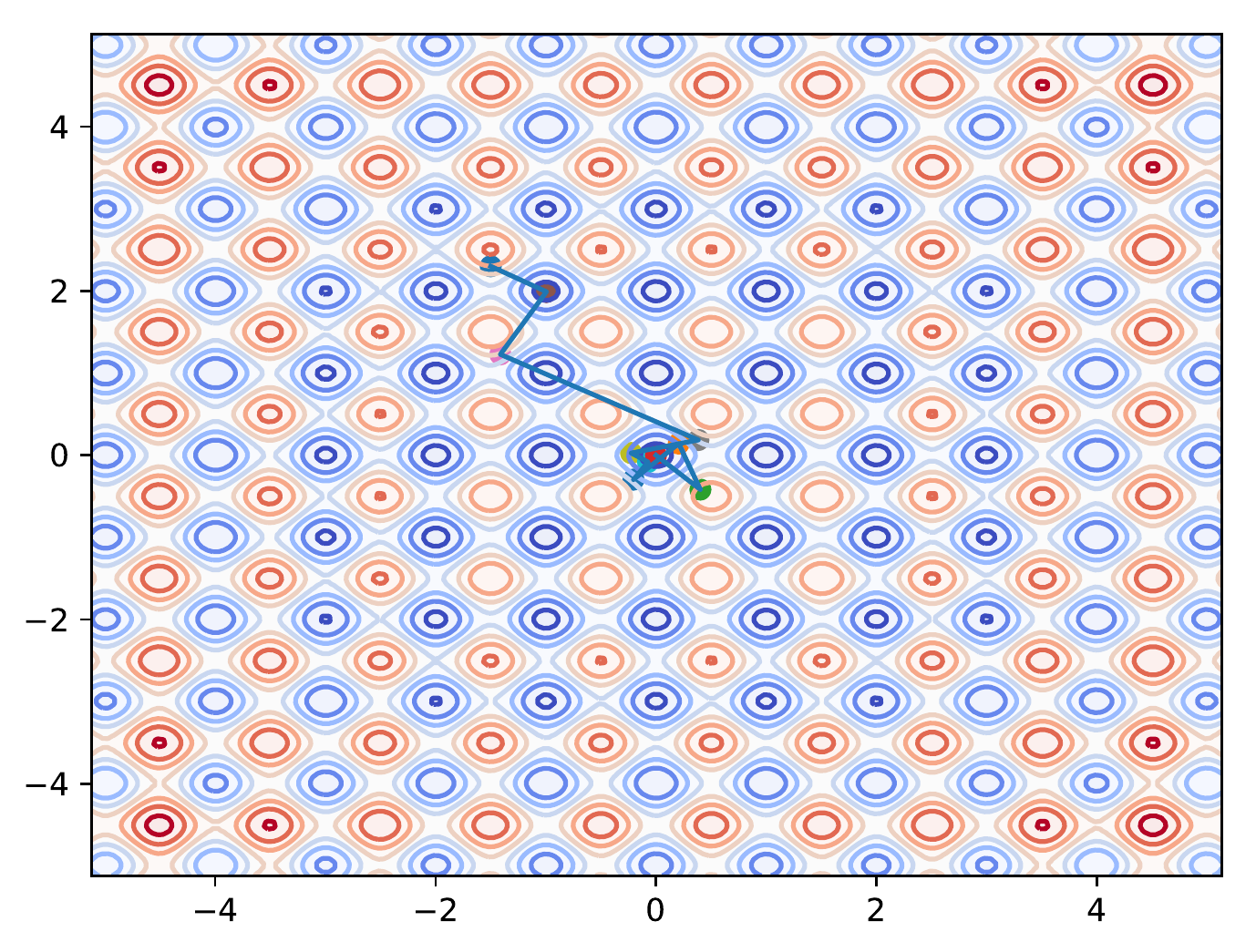}}
\subfigure[Average gradients bound verification in Rastrigin function]{\label{trb}
\includegraphics[width=0.35\linewidth]{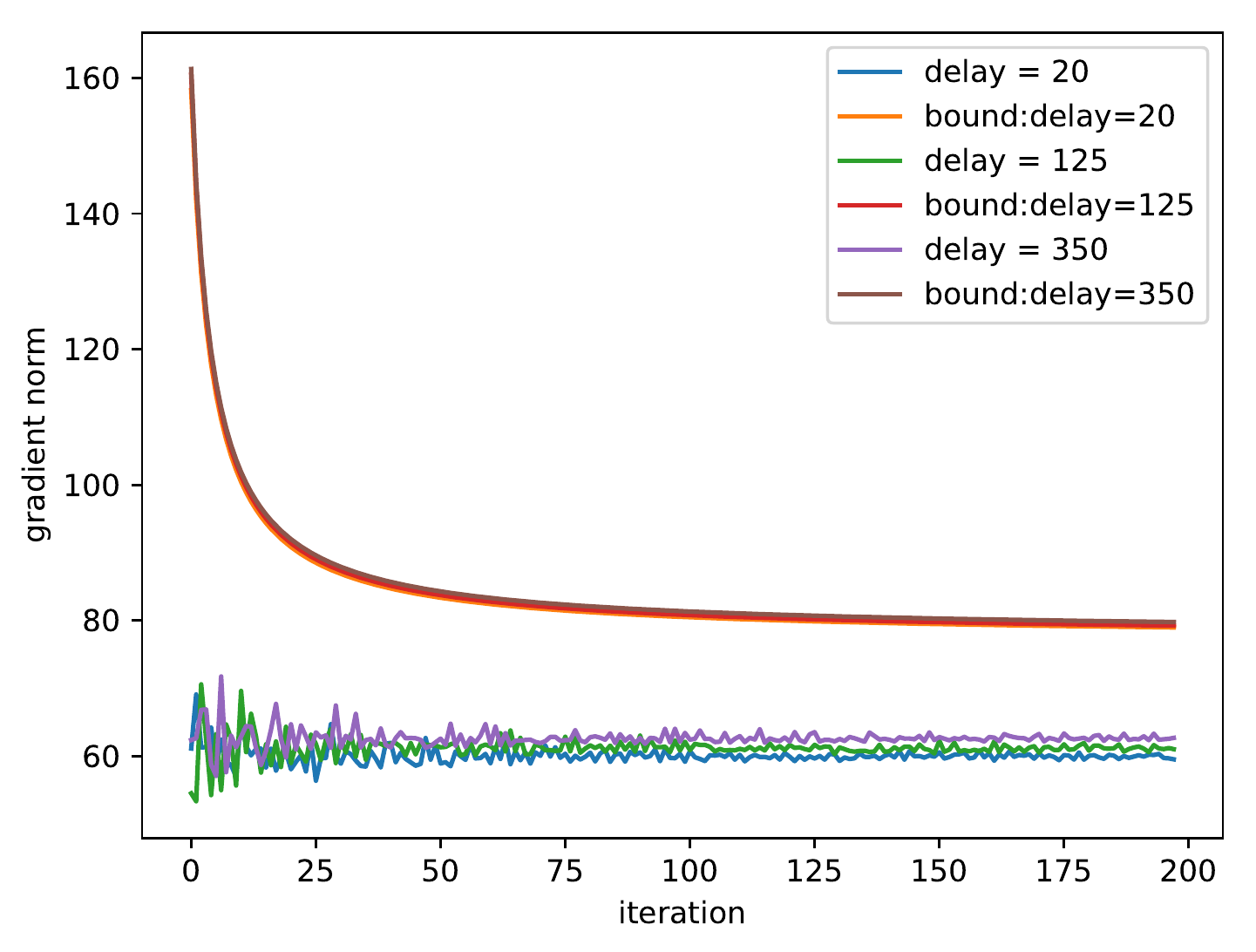}}
\caption{The results of simple functions.}
\label{toyresults}
\end{figure*}

From Fig. \ref{ro} and \ref{rt}, we can view the convergence of our proposed PC-ASGD algorithms. For the bound verification, we take different values of the delay to observe the performances of our theoretical framework. Here, we first find that when delay is large, the squared norm of the gradient is large, which is consistent with our theoretical analysis. 
In Rosenbrock function case, our established theory could describe the tendency of the average gradients square norm and the results are nearly tight asymptotically.
But in Rastrigin function cases, we observe that the differences between different delay are not large such that the bound is not so tight. But when calculating bounds, we find that the bounds for different delays differ mildly, which is consistent along all the empirical results. It also shows the effectiveness of our proposed theoretical analysis.

\section{Conclusion}
\label{secConclu}
This paper presents a novel learning algorithm for distributed deep learning with heterogeneous delay characteristics in agent-communication-network systems. We propose PC-ASGD algorithm consisting of a predicting step, a clipping step, and the corresponding update law for optimizing the positive effects introduced by gradient prediction for reducing the staleness and negative effects caused by the outdated weights. We present theoretical analysis for the convergence rate of the proposed algorithm with constant step size when the objective functions are weakly strongly-convex and nonconvex. The numerical studies show the effectiveness of our proposed algorithms in different distributed systems with delays. In future work, the cases for distributed networks with diverse delays and dynamic topology will be further studied and tested. 

\bibliographystyle{ieeetr}
\bibliography{references}

\begin{thebibliography}{10}

\bibitem{gijzen2013big}
H.~Gijzen, ``Big data for a sustainable future,'' {\em Nature}, vol.~502,
  no.~7469, pp.~38--38, 2013.

\bibitem{wiedemann2019compact}
S.~Wiedemann, K.-R. M{\"u}ller, and W.~Samek, ``Compact and computationally
  efficient representation of deep neural networks,'' {\em IEEE transactions on
  neural networks and learning systems}, vol.~31, no.~3, pp.~772--785, 2019.

\bibitem{lian2017asynchronous}
X.~Lian, W.~Zhang, C.~Zhang, and J.~Liu, ``Asynchronous decentralized parallel
  stochastic gradient descent,'' {\em arXiv: Optimization and Control}, 2017.

\bibitem{hard2018federated}
A.~Hard, C.~Kiddon, D.~Ramage, F.~Beaufays, H.~Eichner, K.~Rao, R.~Mathews, and
  S.~Augenstein, ``Federated learning for mobile keyboard prediction.,'' {\em
  arXiv: Computation and Language}, 2018.

\bibitem{sattler2019robust}
F.~Sattler, S.~Wiedemann, K.-R. M{\"u}ller, and W.~Samek, ``Robust and
  communication-efficient federated learning from non-iid data,'' {\em IEEE
  transactions on neural networks and learning systems}, vol.~31, no.~9,
  pp.~3400--3413, 2019.

\bibitem{ryffel2018a}
T.~Ryffel, A.~Trask, M.~Dahl, B.~Wagner, J.~Mancuso, D.~Rueckert, and
  J.~Passeratpalmbach, ``A generic framework for privacy preserving deep
  learning.,'' {\em arXiv: Learning}, 2018.

\bibitem{deng2020distributionally}
Y.~Deng, M.~M. Kamani, and M.~Mahdavi, ``Distributionally robust federated
  averaging,'' {\em Advances in Neural Information Processing Systems},
  vol.~33, 2020.

\bibitem{strom2015scalable}
N.~Strom, ``Scalable distributed dnn training using commodity gpu cloud
  computing,'' in {\em Sixteenth Annual Conference of the International Speech
  Communication Association}, 2015.

\bibitem{cen2020convergence}
S.~Cen, H.~Zhang, Y.~Chi, W.~Chen, and T.-Y. Liu, ``Convergence of distributed
  stochastic variance reduced methods without sampling extra data,'' {\em IEEE
  Transactions on Signal Processing}, vol.~68, pp.~3976--3989, 2020.

\bibitem{9174890}
F.~Sattler, K.-R. Müller, and W.~Samek, ``Clustered federated learning:
  Model-agnostic distributed multitask optimization under privacy
  constraints,'' {\em IEEE Transactions on Neural Networks and Learning
  Systems}, vol.~32, no.~8, pp.~3710--3722, 2021.

\bibitem{blot2016gossip}
M.~Blot, D.~Picard, M.~Cord, and N.~Thome, ``Gossip training for deep
  learning.,'' {\em arXiv: Computer Vision and Pattern Recognition}, 2016.

\bibitem{even2020asynchrony}
M.~Even, H.~Hendrikx, and L.~Massouli{\'e}, ``Asynchrony and acceleration in
  gossip algorithms,'' {\em arXiv preprint arXiv:2011.02379}, 2020.

\bibitem{li2021consensus}
Z.~Li, B.~Liu, and Z.~Ding, ``Consensus-based cooperative algorithms for
  training over distributed data sets using stochastic gradients,'' {\em IEEE
  Transactions on Neural Networks and Learning Systems}, 2021.

\bibitem{jiang2017collaborative}
Z.~Jiang, A.~Balu, C.~Hegde, and S.~Sarkar, ``Collaborative deep learning in
  fixed topology networks,'' in {\em Advances in Neural Information Processing
  Systems}, pp.~5904--5914, 2017.

\bibitem{liu2019distributed}
B.~Liu, Z.~Ding, and C.~Lv, ``Distributed training for multi-layer neural
  networks by consensus,'' {\em IEEE transactions on neural networks and
  learning systems}, vol.~31, no.~5, pp.~1771--1778, 2019.

\bibitem{chen2016revisiting}
J.~Chen, X.~Pan, R.~Monga, S.~Bengio, and R.~Jozefowicz, ``Revisiting
  distributed synchronous sgd,'' {\em arXiv preprint arXiv:1604.00981}, 2016.

\bibitem{tsianos2012communication}
K.~Tsianos, S.~Lawlor, and M.~G. Rabbat, ``Communication/computation tradeoffs
  in consensus-based distributed optimization,'' in {\em Advances in neural
  information processing systems}, pp.~1943--1951, 2012.

\bibitem{dean2012large}
J.~Dean, G.~S. Corrado, R.~Monga, K.~Chen, and A.~Y. Ng, ``Large scale
  distributed deep networks,'' {\em Advances in neural information processing
  systems}, 2013.

\bibitem{agarwal2011distributed}
A.~Agarwal and J.~C. Duchi, ``Distributed delayed stochastic optimization,''
  {\em arXiv: Optimization and Control}, 2011.

\bibitem{feyzmahdavian2015an}
H.~R. Feyzmahdavian, A.~Aytekin, and M.~Johansson, ``An asynchronous mini-batch
  algorithm for regularized stochastic optimization,'' {\em conference on
  decision and control}, vol.~61, no.~12, pp.~1384--1389, 2015.

\bibitem{recht2011hogwild:}
B.~Recht, C.~Re, S.~Wright, and F.~Niu, ``Hogwild!: A lock-free approach to
  parallelizing stochastic gradient descent,'' {\em Advances in neural
  information processing systems}, vol.~24, pp.~693--701, 2011.

\bibitem{zhuang2021fully}
H.~Zhuang, Y.~Wang, Q.~Liu, and Z.~Lin, ``Fully decoupled neural network
  learning using delayed gradients,'' {\em IEEE Transactions on Neural Networks
  and Learning Systems}, 2021.

\bibitem{zheng2017asynchronous}
S.~Zheng, Q.~Meng, T.~Wang, W.~Chen, N.~Yu, Z.-M. Ma, and T.-Y. Liu,
  ``Asynchronous stochastic gradient descent with delay compensation,'' in {\em
  International Conference on Machine Learning}, pp.~4120--4129, PMLR, 2017.

\bibitem{liang2020asynchrounous}
X.~Liang, A.~M. Javid, M.~Skoglund, and S.~Chatterjee, ``Asynchrounous
  decentralized learning of a neural network,'' in {\em ICASSP 2020-2020 IEEE
  International Conference on Acoustics, Speech and Signal Processing
  (ICASSP)}, pp.~3947--3951, IEEE, 2020.

\bibitem{nair2017wildfire:}
R.~Nair and S.~Gupta, ``Wildfire: approximate synchronization of parameters in
  distributed deep learning,'' {\em Ibm Journal of Research and Development},
  vol.~61, no.~4, p.~7, 2017.

\bibitem{TsianosEfficient}
K.~I. Tsianos and M.~G. Rabbat, ``Efficient distributed online prediction and
  stochastic optimization with approximate distributed averaging,'' {\em IEEE
  Transactions on Signal and Information Processing over Networks}, vol.~2,
  no.~4, pp.~489--506, 2016.

\bibitem{Lan2017Communication}
G.~Lan, S.~Lee, and Y.~Zhou, ``Communication-efficient algorithms for
  decentralized and stochastic optimization,'' {\em Mathematical Programming},
  vol.~180, no.~1, pp.~237--284, 2020.

\bibitem{du2020asynchronous}
Y.~Du, K.~You, and Y.~Mo, ``Asynchronous stochastic gradient descent over
  decentralized datasets,'' in {\em 2020 IEEE 16th International Conference on
  Control \& Automation (ICCA)}, pp.~216--221, IEEE, 2020.

\bibitem{rigazzi2019dc-s3gd:}
A.~Rigazzi, ``Dc-s3gd: Delay-compensated stale-synchronous sgd for large-scale
  decentralized neural network training.,'' {\em arXiv: Learning}, 2019.

\bibitem{2020arXiv200107704Z}
M.~{Zakharov}, ``{Asynchronous Consensus Algorithm},'' {\em arXiv e-prints},
  p.~arXiv:2001.07704, Jan 2020.

\bibitem{venigalla2020adaptive}
A.~Venigalla, A.~Kosson, V.~Chiley, and U.~K{\"o}ster, ``Adaptive braking for
  mitigating gradient delay,'' {\em arXiv preprint arXiv:2007.01397}, 2020.

\bibitem{chen2019communication}
Y.~Chen, X.~Sun, and Y.~Jin, ``Communication-efficient federated deep learning
  with layerwise asynchronous model update and temporally weighted
  aggregation,'' {\em IEEE transactions on neural networks and learning
  systems}, vol.~31, no.~10, pp.~4229--4238, 2019.

\bibitem{2020arXiv200100112A}
S.~{Abbasloo} and H.~J. {Chao}, ``{SharpEdge: An Asynchronous and Core-Agnostic
  Solution to Guarantee Bounded-Delays},'' {\em arXiv e-prints},
  p.~arXiv:2001.00112, Dec 2019.

\bibitem{karimi2016linear}
H.~Karimi, J.~Nutini, and M.~Schmidt, ``Linear convergence of gradient and
  proximal-gradient methods under the polyak-{\l}ojasiewicz condition,'' in
  {\em Joint European Conference on Machine Learning and Knowledge Discovery in
  Databases}, pp.~795--811, Springer, 2016.

\bibitem{carmon2018accelerated}
Y.~Carmon, J.~C. Duchi, O.~Hinder, and A.~Sidford, ``Accelerated methods for
  nonconvex optimization,'' {\em SIAM Journal on Optimization}, vol.~28, no.~2,
  pp.~1751--1772, 2018.

\bibitem{Krizhevsky2009LearningML}
A.~Krizhevsky, ``Learning multiple layers of features from tiny images,'' {\em
  University of Toronto}, 05 2012.

\bibitem{gitcode}
W.~Yang, ``pytorch-classification.''
  \url{https://github.com/bearpaw/pytorch-classification}, 2019.
\newblock Accessed: 2019-01-24.

\bibitem{mishra2006some}
S.~K. Mishra, ``Some new test functions for global optimization and performance
  of repulsive particle swarm method,'' {\em Available at SSRN 926132}, 2006.

\bibitem{friedman2001elements}
J.~Friedman, T.~Hastie, and R.~Tibshirani, {\em The elements of statistical
  learning}, vol.~1.
\newblock Springer series in statistics New York, 2001.

\bibitem{01b8a4abbaba43b48ce43466318a9927}
S.~Becker and Y.~Lecun, ``Improving the convergence of back-propagation
  learning with second-order methods,'' in {\em Proceedings of the 1988
  Connectionist Models Summer School, San Mateo} (D.~Touretzky, G.~Hinton, and
  T.~Sejnowski, eds.), pp.~29--37, Morgan Kaufmann, 1989.

\end{thebibliography}

\clearpage
\appendix
This part presents additional analysis, proof, and experimental results for PC-ASGD.
\section{Additional Analysis}
Before presenting the main results, we introduce some necessary background on the delay compensated gradients. 

\begin{figure}[htbp]
    \centering
    \includegraphics[width=2.4 in]{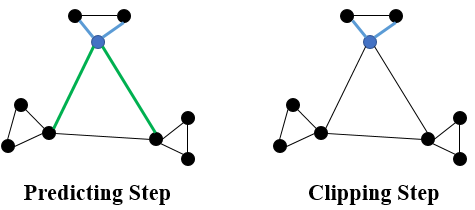}
    \caption{Predicting-Clipping Steps: in the predicting step, blue lines indicate no delay transmission; green lines represent delayed transmission that requires gradient prediction to reduce the stale effect; in the clipping step, the agent selectively drops the delayed information while only receiving information without delay.}
    \label{predic_clip}
\end{figure}

\textbf{Connection Between PC Steps}
As discussed above, PC-ASGD relies upon the two steps to determine the updates for each agent at every time step, as displayed in Fig.~\ref{predic_clip}.  
We first turn to the clipping step (line 7 of Algorithm~\ref{pcasgd}) where all stale information is dropped, which is equivalent to `clipping' the original graph to become a smaller scale of graph. Therefore, between the predicting step and the clipping step, we can observe two static graphs switching alternatively. This also suggests that element values of the mixing matrix $\tilde{W}$ in the clipping step is different from those in the predicting step. In the predicting step (line 6 of Algorithm~\ref{pcasgd}), the agent still requires all the information from its neighbors while asking for gradient prediction from the unreliable neighbors. However, the update is determined by the combination of these two steps in Algorithm~\ref{pcasgd}, which relies on the $\theta$ value to balance the tradeoff.
For simplicity, we set the initialization of each agent $0$. We now turn to the practical variant of PC-ASGD in Algorithm~\ref{pcasgd_1} in Appendix. The condition (line 9) adopted for PC-ASGD is based on the approximate cosine value of the angle between $g_i(x^i_t)$ and $\Delta_{pre}$ (or $\Delta_{clip}$). When the angle between $g_i(x^i_t)$ and $\Delta_{pre}$ (or $\Delta_{clip}$) is smaller, leading to a larger cosine value, the corresponding step should be chosen as it enables a larger descent amount along with the direction of $g_i(x^i_t)$. Hence, with a sequence of graphs and the properly set condition, these two alternating steps are connected to each other, allowing for the convergence.

\textbf{Delay compensated gradient}. We detail how to arrive at Eq.~\ref{delay_comp_gradient}. Specifically, given the outdated weights of agent $k$, $x_{t-\tau}^{k}$, due to the delay equal to $\tau$, by induction, we can obtain for agent $k$
\begin{equation}
\begin{aligned}
x_{t-\tau+1}^{k}=&x_{t-\tau}^{k}-\eta g_k(x_{t-\tau}^{k})\\
=&x_{t-\tau}^{k}-\eta \sum_{r=0}^{0}[g_k(x_{t-\tau}^{k})\\
&+\lambda g_k(x_{t-\tau}^{k})
\odot g_k(x_{t-\tau}^{k})\odot (x_{t-\tau+r}^{i}-x_{t-\tau}^{i})]
\end{aligned}
\end{equation}
\begin{equation}
\begin{split}
x_{t-\tau+2}^{k}&=x_{t-\tau+1}^{k}-\eta g_k(x_{t-\tau+1}^{k})\\
&=x_{t-\tau}^{k}-\eta g_k(x_{t-\tau}^{k})-\eta g_k(x_{t-\tau+1}^{k})\\
\end{split}\notag
\end{equation}
\begin{equation}
\begin{split}
&\approx x_{t-\tau}^{k}-\eta \sum_{r=0}^{1}[g_k(x_{t-\tau}^{k})\\
&+\lambda g_k(x_{t-\tau}^{k})
\odot g_k(x_{t-\tau}^{k})\odot (x_{t-\tau+r}^{i}-x_{t-\tau}^{i})]\\
\cdots
\end{split}
\end{equation}
\begin{equation}
\begin{aligned}
x_{t}^{k}
&\approx x_{t-\tau}^{k}-\eta \sum_{r=0}^{\tau-1}[g_k(x_{t-\tau}^{k})\\
&+\lambda g_k(x_{t-\tau}^{k})\odot g_k(x_{t-\tau}^{k})\odot (x_{t-\tau+r}^{i}-x_{t-\tau}^{i})]\\
\end{aligned}
\end{equation}

\textbf{Compact Form of PC Steps.}
We next briefly discuss how to arrive at the compact form of the predicting and clipping steps for the analysis. For the convenience of analysis, we set the current time step as $t+\tau$ such that line 6 in Algorithm 1 shifts $\tau$ time steps ahead.
Let us start with the predicting step and discussing its associated term 
$\sum_{j \in \mathcal{R}}w_{ij}x_{t+\tau}^{j}+\sum_{k \in \mathcal{R}^c}w_{ik}x_{t}^{k}$. Note that $\mathcal{R}$ includes the agents $i$ itself. Although unreliable neighbors are outdated, in the context, the update for agent $i$ still requires such outdated information, which suggests that the whole graph applies. Additionally, the consensus is performed in parallel with the local computation, so this term boils down to a similar term in the existing consensus-based optimization algorithms in literature. Thus, one can convert the current consensus term for weights to $\sum_{p}w_{ip}x_{t+\tau}^{p}, p\in V$.

Hence, the update law for the predicting step can be rewritten as:
\begin{equation}\label{revised_update_1}
    x_{t+\tau+1}^{i}=\sum_{p}w_{ip}x_{t+\tau}^{p}-\eta(g_k(x_{t+\tau}^{i})+\sum_{k \in \mathcal{R}^c}w_{ik}\sum_{r=0}^{\tau-1}g_k^{dc,r}(x_{t}^{k}))
\end{equation}
One may argue that for those outdated agent $k \in \mathcal{R}^c$, they have no information ahead of time $t$, which is $\tau$ time steps back from the current time. As the graph is undirected and connected, the time scale will not change the connections among agents. Also, for agent $i$, it receives always information from other agents, either the current or the outdated to update its weights. Thus, we have,
\begin{equation}
x_{t+\tau}^{p}= 
\left\{
\begin{array}{rcl}
x_{t+\tau}^{j}   &      & {p=j,j\in \mathcal{R}}\\
x_{t}^{k}   &      & {p=k,k\in \mathcal{R}^c}\\
\end{array}\right. 
\end{equation}

Since the term $\sum_{k \in \mathcal{R}^c}w_{ik}\sum_{r=0}^{\tau-1}g_k^{dc,r}(x_{t}^{k})$ applies to unreliable neighbors only, for the convenience of analysis, we expand it to the whole graph. It means that we establish an expanded graph to cover all of agents by setting some elements in the mixing matrix $\underline{W}'\in\mathbb{R}^{N\times N}$ equal to $0$, but keeping the same connections as in $\underline{W}$.
Then Eq.~\ref{revised_update_1} can be modified as
\begin{equation}
x_{t+\tau+1}^{i}=\sum_{p}w_{ip}x_{t+\tau}^{p}-\eta(g_k(x_{t+\tau}^{i})+\sum_{q}w'_{iq}\sum_{r=0}^{\tau-1}g_k^{dc,r}(x_{t}^{q}))
\end{equation}
where
\begin{equation}
\begin{footnotesize}
w'_{iq}=\left\{
\begin{array}{rcl}
w_{ik}   &      & {if\,\ q=k,k \in \mathcal{R}^c}\\
0  &      & {if\,\ q\in \mathcal{R}}\\
\end{array}\right. 
\end{footnotesize}
\end{equation}
Thus, we can know via the above setting that $\underline{W}'$ is at least a row stochastic matrix.
We rewrite the update law into a compact form such that 
\begin{equation}\label{vector_pre}
\begin{footnotesize}
    \mathbf{x}_{t+\tau+1}=W \mathbf{x}_{t+\tau}-\eta(\mathbf{g}(\mathbf{x}_{t+\tau})+\sum_{r=0}^{\tau-1}W'\mathbf{g}^{dc,r}(\mathbf{x}_{t})).
\end{footnotesize}
\end{equation}
where $W=\underline{W}\otimes I_{d\times d}$ and $W'=\underline{W}'\otimes I_{d\times d}$.
Similarly, we rewrite the clipping steps in a vector form as follows:
\begin{equation}\label{vector_clip}
\begin{footnotesize}
    \mathbf{x}_{t+\tau+1} = \tilde{W}\mathbf{x}_{t+\tau} - \eta\mathbf{g}(\mathbf{x}_{t+\tau})
\end{footnotesize}
\end{equation}
where $\tilde{W}=\tilde{\underline{W}}\otimes I_{d\times d}$.
We are now ready to give the generalized step
\begin{equation}\label{g_pcasgd}
\begin{split}
    \mathbf{x}_{t+\tau+1}=\mathcal{W}_{t+\tau}\mathbf{x}_{t+\tau}-\eta(\mathbf{g}(\mathbf{x}_{t+\tau})+\theta_{t+\tau}\sum_{r=0}^{\tau-1}W'\mathbf{g}^{dc,r}(\mathbf{x}_{t})),
\end{split}
\end{equation}
where $\mathcal{W}_{t+\tau}$ is denoted as $\theta_{t+\tau}W+(1-\theta_{t+\tau})\tilde{W}$ throughout the rest of the analysis. Though the original graphs corresponding to the predicting and clipping steps are static, the equivalent graph $\mathcal{W}_{t+\tau}$ has become time-varying due to the time-varying $\theta$ value.

\textbf{Approximate Hessian Matrix}.
Based on the update law, we have known that the key part of PC-ASGD is the delay compensated gradients using Taylor expansion and Hessian approximation. Therefore, the Taylor expansion of the stochastic gradient $\mathbf{g}(\mathbf{x}_{t+\tau})$ at $\mathbf{x}_{t}$ can be written as follows:
\begin{equation}
    \mathbf{g}(\mathbf{x}_{t+\tau})=\mathbf{g}(\mathbf{x}_{t})+\nabla \mathbf{g}(\mathbf{x}_{t})(\mathbf{x}_{t+\tau}-\mathbf{x}_{t})+O((\mathbf{x}_{t+\tau}-\mathbf{x}_t)^2)I,
\end{equation}

where $\nabla \mathbf{g}$ denotes the matrix with the element $\nabla g_{ij}=\frac{\partial F}{\partial x^{i}\partial x^{j}}$ for all $i,j \in V$.

In most asynchronous SGD works, they used the zero-order item in Taylor expansion as its approximation to $\mathbf{g}(\mathbf{x}_{t+\tau})$ by ignoring the higher order term. Following from~\cite{zheng2017asynchronous}, we have 
\begin{equation}
    \mathbf{g}(\mathbf{x}_{t+\tau})\approx \mathbf{g}(\mathbf{x}_t)+\nabla \mathbf{g}(\mathbf{x}_{t})(\mathbf{x}_{t+\tau}-\mathbf{x}_{t}),
\end{equation}

Directly adopting the above equation would be difficult in practice since $\nabla \mathbf{g}(\mathbf{x}_t)$ is generically computationally intractable when the model is very large, such as deep neural networks. To 
make the delay compensated gradients in PC-ASGD technically feasible, we apply approximation techniques for the Hessian matrix.
We first use $O(\mathbf{x}_t)$ to denote the outer product matrix of the gradient at $\mathbf{x}_t$, i.e., 
\begin{equation}
    O(\mathbf{x}_t)=(\frac{\partial}{\partial \mathbf{x}_t}F(\mathbf{x}_t))(\frac{\partial}{\partial \mathbf{x}_t}F(\mathbf{x}_t))^{T}
\end{equation}

When the objective functions are the cross-entropy loss like, or negative log-likelihood forms, the outer product of the gradient is an asymptotically unbiased estimation of the Hessian, according to the two equivalent methods to calculate the Fisher information matrix~\cite{friedman2001elements}. That is,
\begin{equation}
    \epsilon_t=\mathbb{E}[\|O(\mathbf{x}_t)-H(\mathbf{x}_t)\|]\rightarrow0,\,\,\,\ t\rightarrow 0
\end{equation}
where $H(\mathbf{x}_t)$ is the Hessian matrix of $F$ at point $\mathbf{x}_t$.

The above equivalence relies on assumptions that the underlying distribution equals the model distribution with parameter $\mathbf{x}^{*}$ and that the training model $\mathbf{x}_t$
asymptotically converges to the (globally or locally) optimal model $\mathbf{x}^{*}$. According to the universal approximation theorem for DNN and some recent results on the optimality of the local optimal, such assumptions are technically reasonable. As the above equivalence was only developed by the negative log-likelihood form, that may not be applicable when we use PC-ASGD for the mean square error form, such as some time-series prediction with LSTM networks. Therefore, we introduce one assumption on the top of such an equivalence as follows,
\begin{equation}
\mathbb{E}[\|O(\mathbf{x}_t)-H(\mathbf{x}_t)\|]\leq \epsilon \,\,\,\ \exists \epsilon>0
\end{equation}

which primarily eliminates the computational complexity when directly calculating $H(\mathbf{x}_t)$. Another concern would be the large variance probably caused by $O(\mathbf{x}_t)$, though it is an unbiased estimation of $H(\mathbf{x}_t)$. Similar to~\cite{zheng2017asynchronous}, we introduce a new approximator $\lambda O(\mathbf{x}_t) \triangleq \lambda (\frac{\partial}{\partial \mathbf{x}_t}F(\mathbf{x}_t))(\frac{\partial}{\partial \mathbf{x}_t}F(\mathbf{x}_t))^{T}$. The authors in~\cite{zheng2017asynchronous} have proved that $\lambda O(\mathbf{x}_t)$ is able to lead to smaller variance during training. Thus we refer interested readers to~\cite{zheng2017asynchronous} for more details.

To reduce the storage of the approxiamtor $\lambda O(\mathbf{x}_t)$, one widely-used diagonalization trick is adopted~\cite{01b8a4abbaba43b48ce43466318a9927}. Hence, in the update law for PC-ASGD, we can see in the delay compensated gradient involving $\lambda g(\mathbf{x}_t) \odot \lambda g(\mathbf{x}_t)$. By denoting the diagonalized approximator as $Diag(\lambda O(\mathbf{x}_t))$, the following relationship is obtained:
 \begin{equation}
    Diag(\lambda O(\mathbf{x}_t)) = \lambda g(\mathbf{x}_t) \odot \lambda g(\mathbf{x}_t)
 \end{equation}
 
 However, for analysis, when we apply diagonalization to $H(\mathbf{x}_t)$, it could cause diagonalization error such that we assume that the error is upper bounded by a constant $\epsilon_D>0$, i.e.,
 \begin{equation}
     \|Diag(H(\mathbf{x}_t))-H(\mathbf{x}_t)\|\leq \epsilon_D
 \end{equation}
\section{Additional Proof} \label{secproof}
For completeness, when presenting proof, we re-present statements for all lemmas and theorems.

\noindent $\textbf{Lemma 2}$: The iterates generated by PC-ASGD satisfy $\forall t\geq 0$, and $\tau \geq 2$:
\begin{equation}
\begin{aligned}
\mathbf{x}_{t+\tau}=&\prod^{t+\tau-1}_{v=0}\mathcal{W}_{t+\tau-1-v}\mathbf{x}_{0}-\eta\sum_{s=0}^{t+\tau-1}\prod^{t+\tau-1}_{v=s+1}\mathcal{W}_{t+\tau+s-v}\mathbf{g}(\mathbf{x}_{s})\\
&-\eta\sum_{s=t}^{t+\tau-1}\prod^{t+\tau-1}_{v=s+1}\theta_{s+1}\mathcal{W}_{t+\tau+s-v}\sum_{r=0}^{\tau-2}W'\mathbf{g}(\mathbf{x}_{s+1}).
\end{aligned}
\end{equation}
\begin{proof}
Based on the vector form of the update law, we obtain
\begin{equation}
\mathbf{x}_{t+\tau}=\mathcal{W}_{t+\tau-1} \mathbf{x}_{t+\tau-1}-\eta(\mathbf{g}(\mathbf{x}_{t+\tau-1})+\theta_{t+\tau-1}\sum_{r=0}^{\tau-2}W'\mathbf{g}^{dc,r}(\mathbf{x}_{t}))
\end{equation}
With the above equation, it can be observed that $\mathbf{x}_{t+\tau}$ is a function with respect to $\mathbf{x}_{t}$, which contains all of agents. This suggests that by $\mathbf{x}_{t}$, there were no delay compensated gradients, while after $\mathbf{x}_{t+1}$, the unreliable neighbors need the delay compensated gradients due to delay. Hence, applying the above equation from 0 to $t+\tau-1$ yields the desired result.
\end{proof}
\textbf{Bounded (stochastic) gradient}: As $\mathbb{E}[\|\mathbf{g}(\mathbf{x})\|^2]\leq G^2$ and $\mathbb{E}[\mathbf{g}(\mathbf{x})]=\nabla F(\mathbf{x})$, one can get that $\|\nabla F(\mathbf{x})\|=\|\mathbb{E}[\mathbf{g}(\mathbf{x})]\|\leq\mathbb{E}[\|\mathbf{g}(\mathbf{x})\|]=\sqrt{(\mathbb{E}[\|\mathbf{g}(\mathbf{x})\|])^2}\leq\sqrt{\mathbb{E}[\|\mathbf{g}(\mathbf{x})\|^2]}=G$.

\noindent $\textbf{Lemma 1}$: Let Assumptions 2 and 3 hold. Assume that the delay compensated gradients are uniformly bounded, i.e., there exists a scalar $B>0$, such that 
\begin{equation}
    \|\mathbf{g}^{dc,r}(\mathbf{x}_{t})\|\leq B,\,\,\ \forall t\geq 0 \,\ and \,\ 0\leq r\leq\tau-1,
\end{equation}
Then for all $i\in V$ and $t\geq0$, $\exists \eta > 0$, we have
\begin{equation}
\mathbb{E}[\|x_{t}^{i}-y_{t}\|]\leq 
\eta\frac{ G+(\tau-1)B\theta_m}{1-\delta_2},
\end{equation}
where $\theta_m=\text{max}\{\theta_{s+1}\}^{t+\tau-1}_{s=t}$, $\delta_2=\text{max}\{\theta_se_2+(1-\theta_s)\tilde{e}_2\}^{t+\tau-1}_{s=0}<1$, where $e_{2}:=e_{2}(W) < 1$ and $\tilde{e}_{2}:=e_{2}(\tilde{W}) < 1$.

\begin{proof}
Since
\begin{equation}
\begin{aligned}
    \|x_{t+\tau}^{i}-y_{t+\tau}\|&\leq \|\mathbf{x}_{t+\tau}-y_{t+\tau}\mathbf{1}\|\\
    &=\|\mathbf{x}_{t+\tau}-\frac{1}{N}\mathbf{1}^{T}\mathbf{x}_{t+\tau}\mathbf{1}\|\\
    &=\|\mathbf{x}_{t+\tau}-\frac{1}{N}\mathbf{1}\mathbf{1}^{T}\mathbf{x}_{t+\tau}\|\\
    &=\|(I-\frac{1}{N}\mathbf{1}\mathbf{1}^{T})\mathbf{x}_{t+\tau}\|,
\end{aligned}
\end{equation}
where $\mathbf{1}$ is the column vector with entries all being 1.
According to Assumption 2, we have $\frac{1}{N}\mathbf{1}\mathbf{1}^{T}\mathcal{W}=\frac{1}{N}\mathbf{1}\mathbf{1}^{T}$. Hence, by induction, setting $\mathbf{x}_{0}=0$, and Lemma 1, the following relationship can be obtained
\begin{equation}
\begin{aligned}
&\|\mathbf{x}_{t+\tau}-y_{t+\tau}\mathbf{1}\|=\\&\eta\|\sum_{s=0}^{t+\tau-1}(\prod^{t+\tau-1}_{v=s+1}\mathcal{W}_{t+\tau+s-v}-\frac{1}{N}\mathbf{1}\mathbf{1}^{T})\mathbf{g}(\mathbf{x}_{s})\\&+\sum_{s=t}^{t+\tau-1}(\prod^{t+\tau-1}_{v=s+1}\mathcal{W}_{t+\tau+s-v}-\frac{1}{N}\mathbf{1}\mathbf{1}^{T})\\&\theta_{s+1}\sum_{r=0}^{\tau-2}W'\mathbf{g}^{dc,r}(\mathbf{x}_{t})\|\\
\leq& \eta\sum_{s=0}^{t+\tau-1}\|\prod^{t+\tau-1}_{v=s+1}\mathcal{W}_{t+\tau+s-v}-\frac{1}{N}\mathbf{1}\mathbf{1}^{T}\|\|\mathbf{g}(\mathbf{x}_s)\|\\&+\eta\sum_{s=t}^{t+\tau-1}\|\prod^{t+\tau-1}_{v=s+1}\mathcal{W}_{t+\tau+s-v}-\frac{1}{N}\mathbf{1}\mathbf{1}^{T}\|\\&\|\theta_{s+1}\sum_{r=0}^{\tau-2}W'\mathbf{g}^{dc,r}(\mathbf{x}_{t})\|\\
\leq& \eta G\sum_{s=0}^{t+\tau-1}\delta_2^{t+\tau-1-s}+\eta\sum_{s=t}^{t+\tau-1}\delta_2^{t+\tau-1-s}\theta_{s+1}(\tau-1)B\\
\leq& \eta G\frac{1}{1-\delta_2}+\eta(\tau-1)B\theta_m\frac{\delta^t_2-\delta_2^{t+\tau-1}}{1-\delta_2}\\&\leq\eta\frac{G+(\tau-1)B\theta_m}{1-\delta_2}.
\end{aligned}
\end{equation}
The second inequality follows from the Triangle inequality and Cauthy-Schwartz inequality and the third inequality follows from Assumption~\ref{assump2} and that the matrix $\frac{1}{N}\mathbf{1}\mathbf{1}^{T}$ is the projection of $\mathcal{W}$ onto the eigenspace associated with the eigenvalue equal to $1$. The last inequality follows from the property of geometric sequence. The proof is completed by replacing $t+\tau$ with $t$ on the left hand side.
\end{proof}

To prove the main results, we present several auxiliary lemmas first. We define
\begin{equation}
\begin{aligned}
\mathcal{G}^h(\mathbf{x}_t)=\sum_{r=0}^{\tau-1}\mathbf{g}(\mathbf{x}_{t+r})+H(\mathbf{x}_t)(\mathbf{v}_{t+r}-\mathbf{x}_t)\\
\nabla \mathcal{F}^h(\mathbf{x}_t)=\sum_{r=0}^{\tau-1}\nabla F(\mathbf{x}_{t+r})+\mathbb{E}[H(\mathbf{x}_t)(\mathbf{v}_{t+r}-\mathbf{x}_t)]
\end{aligned}
\end{equation}
which are the incrementally delay compensated gradient and its expectation, respectively. It can be observed that $\mathcal{G}^h(\mathbf{x}_t)$ is the unbiased estimator of $\nabla \mathcal{F}^h(\mathbf{x}_t)$. It should be noted that $H(\mathbf{x}_t)=\nabla \mathbf{g}(\mathbf{x}_t)$.
Let $\mathbf{v}_{t+\tau}=\mathcal{W}_{t+\tau} \mathbf{x}_{t+\tau}$. We next present a lemma to upper bound $\|\nabla F(\mathbf{v}_{t+r})-\nabla \mathcal{F}^{h,r}(\mathbf{x}_t)\|$, where $\nabla \mathcal{F}^{h,r}(\mathbf{x}_t)=\nabla F(\mathbf{x}_{t+r})+\mathbb{E}[H(\mathbf{x}_t)(\mathbf{v}_{t+r}-\mathbf{x}_t)]$.

\noindent $\textbf{Lemma 3}$: Let Assumptions 1,2 and 3 hold. Assume that $\nabla F(\mathbf{x}_t)$ is $\xi_m$-smooth. For $t\geq0$, the iterates generated by PC-ASGD satisfies the following relationship, when $r\geq1$
\begin{equation}
\begin{aligned}
&\|\nabla F(\mathbf{v}_{t+r})-\nabla \mathcal{F}^{h,r}(\mathbf{x}_t)\|\leq\frac{\xi_m}{2}\eta^2(\frac{2G+(r-1)B\theta_m}{1-\delta_2})^{2};
\end{aligned}
\end{equation}
when $r=0$, we have
\begin{equation}
\begin{aligned}
&\|\nabla F(\mathbf{v}_{t})-\nabla F(\mathbf{x}_t))\|\leq 2\gamma_m\frac{\eta(G+(\tau-1)B\theta_m)}{1-\delta_2}.
\end{aligned}
\end{equation}

\begin{proof}
By the smoothness condition for $\nabla F(\mathbf{x})$, we have
\begin{equation}
\begin{aligned}
\|\nabla F(\mathbf{v}_{t+r})-\nabla \mathcal{F}^{h,r}(\mathbf{x}_t)\|&\leq\frac{\xi_m}{2}\|\mathbf{v}_{t+r}-\mathbf{x}_t\|^2\\&
\leq\frac{\xi_m}{2}\|\mathbf{x}_{t+r}-\mathbf{x}_t\|^2
\end{aligned}
\end{equation}
Let $\mathbf{\Delta}_{t+r}=\mathbf{x}_{t+r}-\mathbf{x}_t$. Thus, based on Lemma 1, we have
\begin{equation}
\begin{split}
\mathbf{x}_{t+r}&=\prod_{v=t}^{t+r-1}\mathcal{W}_{t+r-1-v}\mathbf{x}_{t}-\eta\sum_{s=t}^{t+r-1}\prod_{v=s+1}^{t+r-1}\mathcal{W}_{t+r+s-v}\mathbf{g}(\mathbf{x}_s)\\&-\eta\sum_{s=t}^{t+r-1}\prod_{v=s+1}^{t+r-1}\mathcal{W}_{t+s+r-v}\sum_{z=0}^{r-2}\theta_{s+1}W'\mathbf{g}^{dc,z}(\mathbf{x}_{s+1-r})
\end{split}
\end{equation}
Hence, we can obtain
\begin{equation}
\begin{split}
&\|\mathbf{\Delta}_{t+r}\|^2
\\&=\|(\prod_{v=t}^{t+r-1}\mathcal{W}_{t+r-1-v}-I)\mathbf{x}_{t}\\&-\eta\sum_{s=t}^{t+r-1}\prod_{v=s+1}^{t+r-1}\mathcal{W}_{t+r+s-v}\mathbf{g}(\mathbf{x}_s)\\&-\eta\sum_{s=t}^{t+r-1}\prod_{v=s+1}^{t+r-1}\mathcal{W}_{t+s+r-v}\sum_{z=0}^{r-2}\theta_{s+1}W'\mathbf{g}^{dc,z}(\mathbf{x}_{s+1-r})\|^2
\end{split}
\end{equation}
Due to $\mathbf{x}_0=0$ and no delay compensated gradients before time step $t$, we can obtain
\begin{equation}
\begin{split}
&\|\mathbf{\Delta}_{t+r}\|^2\\
=&\|-\eta\sum_{s=0}^{t+r-1}\prod_{v=s+1}^{t+r-1}\mathcal{W}_{t+r+s-v}\mathbf{g}(\mathbf{x}_s)\\&-\eta\sum_{s=t}^{t+r-1}\prod_{v=s+1}^{t+r-1}\mathcal{W}_{t+s+r-v}\sum_{z=0}^{r-2}\theta_{s+1}W'\mathbf{g}^{dc,z}(\mathbf{x}_{s+1-r})\\&+\eta\sum^t_{s=0}\prod_{v=s}^t\mathcal{W}_{t+s-v}\mathbf{g}(\mathbf{x}_s)\|^2 \\
\leq&\eta^2(\|\sum_{s=0}^{t+r-1}\prod_{v=s+1}^{t+r-1}\mathcal{W}_{t+r+s-v}\mathbf{g}(\mathbf{x}_s)\|\\&+\|\sum_{s=t}^{t+r-1}\prod_{v=s+1}^{t+r-1}\mathcal{W}_{t+s+r-v}\sum_{z=0}^{r-2}\theta_{s+1}W'\mathbf{g}^{dc,z}(\mathbf{x}_{s+1-r})\|\\&+\|\sum^t_{s=0}\prod_{v=s}^t\mathcal{W}_{t+s-v}\mathbf{g}(\mathbf{x}_s)\|)^{2}\\
\leq& \eta^2(\sum_{s=0}^{t+r-1}\|\prod^{t+r-1}_{v=s+1}\mathcal{W}_{t+r+s-v}\mathbf{g}(\mathbf{x}_s)\|\\&+\sum_{s=t}^{t+r-1}\|\prod_{v=s+1}^{t+r-1}\mathcal{W}_{t+s+r-v}\sum_{z=0}^{r-2}\theta_{s+1}W'\mathbf{g}^{dc,z}(\mathbf{x}_{s+1-r})\|\\&+\sum^t_{s=0}\|\prod_{v=s}^t\mathcal{W}_{t+s-v}\mathbf{g}(\mathbf{x}_s)\|)^2 \\
\leq&\eta^2(\sum_{s=0}^{t+r-1}\prod^{t+r-1}_{v=s+1}\|\mathcal{W}_{t+r+s-v}\|\|\mathbf{g}(\mathbf{x}_s)\|\\&+\sum_{s=t}^{t+r-1}\prod_{v=s+1}^{t+r-1}\|\mathcal{W}_{t+s+r-v}\|\|\sum_{z=0}^{r-2}\theta_{s+1}W'\mathbf{g}^{dc,z}(\mathbf{x}_{s+1-r})\|\\&+\sum^t_{s=0}\prod_{v=s}^t\|\mathcal{W}_{t+s-v}\|\|\mathbf{g}(\mathbf{x}_s)\|)^2 \\
\leq&\eta^2(\frac{2G}{1-\delta_2}+\frac{1}{1-\delta_2}B(r-1)\theta_m)^2\\
\leq&\eta^2(\frac{2G+\theta_m(r-1)B}{1-\delta_2})^{2}
\end{split}
\end{equation}
The first inequality follows from the Triangle inequality. The second inequality follows from the Jensen inequality. The third inequality follows from the Cauthy-Schwartz inequality and the submultiplicative matrix norm applied to stochastic matrices. The fourth inequality follows from the Assumption~\ref{assump2} and bounded gradient. 
We have observed that this holds when $r\geq1$.
While $r=0$ enables $\|\nabla F(\mathbf{v}_{t+r})-\mathcal{F}^{h,r}(\mathbf{x}_t)\|$ to degenerate to $\|\nabla F(\mathbf{v}_{t})-\nabla F(\mathbf{x}_t))\|$ based on the definition of $\mathcal{F}^{h}(\mathbf{x}_t)$. Using the smoothness condition of $F(\mathbf{x})$, we can immediately obtain
\begin{equation}
\begin{aligned}
&\|\nabla F(\mathbf{v}_{t})-\nabla F(\mathbf{x}_t))\|\leq 2\gamma_m\eta\frac{G+(\tau-1)B\theta_m}{1-\delta_2}.
\end{aligned}
\end{equation}
The proof is completed.
\end{proof}

\noindent $\textbf{Lemma 4}$: Let Assumptions 1, 2 and 3 hold. Assume that the delay compensated gradients are uniformly bounded, i.e., there exists a scalar $B>0$ such that 
\begin{equation}
    \|\mathbf{g}^{dc,r}(\mathbf{x}_{t})\|\leq B,\,\,\ \forall t\geq 0 \,\ and \,\ 0\leq r\leq\tau-1,
\end{equation}
Then for the iterates generated by PC-ASGD, $\exists \eta>0$, they satisfy
\begin{equation}
\begin{aligned}
&\|\mathbb{E}[\mathcal{G}^h(\mathbf{x}_t)]-\sum_{r=0}^{\tau-1}W'\mathbf{g}^{dc,r}(\mathbf{x}_t)\|\\
&\leq\sum_{r=1}^{\tau-1}(\gamma_m+\epsilon_D+\epsilon
+(1-\lambda)G^2)\eta\frac{2G+(r-1)B\theta_m}{1-\delta_2}+\tau\sigma
\end{aligned}
\end{equation}
\begin{proof}
Based on the definition of $\mathbb{E}\mathcal{G}^h(\mathbf{x}_t)$, we have 
\begin{equation}
\begin{split}
&\|\mathbb{E}[\mathcal{G}^h(\mathbf{x}_t)]-\sum_{r=0}^{\tau-1}W'\mathbf{g}^{dc,r}(\mathbf{x}_t)\|\\
=&\|\mathbb{E}[\sum_{r=0}^{\tau-1}\mathbf{g}(\mathbf{x}_{t+r})+\sum_{r=0}^{\tau-1}H(\mathbf{x}_t)(\mathbf{x}_{t+r}-\mathbf{x}_t)]\\&-\sum_{r=0}^{\tau-1}W'\mathbf{g}^{dc,r}(\mathbf{x}_t)\|\\
=&\|\mathbb{E}[\mathcal{G}^{h,r=0}(\mathbf{x}_t)]-W'\mathbf{g}^{dc,r=0}(\mathbf{x}_t)\\&+\mathbb{E}[\mathcal{G}^{h,r=1}(\mathbf{x}_t)]-W'\mathbf{g}^{dc,r=1}(\mathbf{x}_t)+...\\&+\mathbb{E}[\mathcal{G}^{h,r=\tau-1}(\mathbf{x}_t)]-W'\mathbf{g}^{dc,r=\tau-1}(\mathbf{x}_t)\|\\
\end{split}\notag
\end{equation}
\begin{equation}
\begin{split}
\leq&\|\mathbb{E}[\mathcal{G}^{h,r=0}(\mathbf{x}_t)]-W'\mathbf{g}^{dc,r=0}(\mathbf{x}_t)\|+\|\mathbb{E}[\mathcal{G}^{h,r=1}(\mathbf{x}_t)]\\&-W'\mathbf{g}^{dc,r=1}(\mathbf{x}_t)\|+...+\|\mathbb{E}[\mathcal{G}^{h,r=\tau-1}(\mathbf{x}_t)]\\&-W'\mathbf{g}^{dc,r=\tau-1}(\mathbf{x}_t)\|
\end{split}
\end{equation}
The last inequality follows from the Triangle inequality.
Now let us discuss $\|\mathbb{E}\mathcal{G}^{h,r}(\mathbf{x}_t)-W'\mathbf{g}^{dc,r}(\mathbf{x}_t)\|$. The following analysis is for cases where $r\geq1$. We give a brief analysis for case in which $r=0$ subsequently.
\begin{equation}
\begin{split}
&\|\mathbb{E}[\mathcal{G}^h(\mathbf{x}_t)]-W'\mathbf{g}^{dc,r}(\mathbf{x}_t)\|
\\&=\|\mathbb{E}[\mathbf{g}(\mathbf{x}_{t+r})+H(\mathbf{x}_t)(\mathbf{x}_{t+r}-\mathbf{x}_t)]-\\&W'[\mathbf{g}(\mathbf{x}_t)+\lambda \mathbf{g}(\mathbf{x}_t)\odot \mathbf{g}(\mathbf{x}_t)\odot(\mathbf{x}_{t+r}-\mathbf{x}_t)]\| \\
=&\|\nabla F(\mathbf{x}_{t+r})-W'\mathbf{g}(\mathbf{x}_t) \\&+[H(\mathbf{x}_t)-\lambda W' \mathbf{g}(\mathbf{x}_t)\odot \mathbf{g}(\mathbf{x}_t)](\mathbf{x}_{t+r}-\mathbf{x}_t)\| \\
\leq& \|\nabla F(\mathbf{x}_{t+r})-W'\mathbf{g}(\mathbf{x}_t)\|\\&+\|[H(\mathbf{x}_t)-\lambda W' \mathbf{g}(\mathbf{x}_t)\odot \mathbf{g}(\mathbf{x}_t)](\mathbf{x}_{t+r}-\mathbf{x}_t)\| \\
\leq& \|\nabla F(\mathbf{x}_{t+r})-W'\mathbf{g}(\mathbf{x}_t)\|\\&+\|[H(\mathbf{x}_t)-\lambda W' \mathbf{g}(\mathbf{x}_t)\odot \mathbf{g}(\mathbf{x}_t)+\mathbf{g}(\mathbf{x}_t)\odot \mathbf{g}(\mathbf{x}_t) \nonumber \\
&-\mathbf{g}(\mathbf{x}_t)\odot \mathbf{g}(\mathbf{x}_t)-Diag(H(\mathbf{x}_t))+Diag(H(\mathbf{x}_t))]\\&(\mathbf{x}_{t+r}-\mathbf{x}_t)\| \nonumber \\
\leq& \|\nabla F(\mathbf{x}_{t+r})-W'\mathbf{g}(\mathbf{x}_t)\|
\\&+\|\mathbf{x}_{t+r}-\mathbf{x}_t\|\|(\lambda W' \mathbf{g}(\mathbf{x}_t)\odot \mathbf{g}(\mathbf{x}_t) -\mathbf{g}(\mathbf{x}_t)\odot \mathbf{g}(\mathbf{x}_t))\\
&+(\mathbf{g}(\mathbf{x}_t)\odot \mathbf{g}(\mathbf{x}_t)-Diag(H(\mathbf{x}_t))) \nonumber \\&+(Diag(H(\mathbf{x}_t))-H(\mathbf{x}_t))\|\\
\leq& \|\nabla F(\mathbf{x}_{t+r})-W'\mathbf{g}(\mathbf{x}_t)\|
\\
\end{split}
\end{equation}
\begin{equation}
\begin{split}
&+\|\mathbf{x}_{t+r}-\mathbf{x}_t\|(\|\lambda W' \mathbf{g}(\mathbf{x}_t)\odot \mathbf{g}(\mathbf{x}_t) -\mathbf{g}(\mathbf{x}_t)\odot \mathbf{g}(\mathbf{x}_t)\|\\
&+\|\mathbf{g}(\mathbf{x}_t)\odot \mathbf{g}(\mathbf{x}_t)-Diag(H(\mathbf{x}_t))\| \nonumber \\&+\|Diag(H(\mathbf{x}_t))-H(\mathbf{x}_t)\|)
\end{split}
\end{equation}
The third inequality follows from Cauthy-Schwarz inequality while the last one follows from the Triangle inequality.
It should be noted that when we combine $H(\mathbf{x}_t)(\mathbf{x}_{t+r}-\mathbf{x}_t)$ and $\lambda W' \mathbf{g}(\mathbf{x}_t)\odot \mathbf{g}(\mathbf{x}_t)\odot(\mathbf{x}_{t+r}-\mathbf{x}_t)$, we follow the update law. Since in a rigorously mathematical sense, $\mathbf{g}(\mathbf{x}_t)\odot \mathbf{g}(\mathbf{x}_t)$ should be $\mathbf{g}(\mathbf{x}_t)\mathbf{g}(\mathbf{x}_t)^{T}$. However, for reducing the computational complexity when implementing the algorithm, as discussed above, we have made the approximation and diagonalization trick. Hence, we assume that $H(\mathbf{x}_t)-\lambda W' \mathbf{g}(\mathbf{x}_t)\odot \mathbf{g}(\mathbf{x}_t)$ can hold for simplicity and convenience.

Then we discuss $\mathbb{E}[\|\nabla F(\mathbf{x}_{t+r})-W'\mathbf{g}(\mathbf{x}_t)\|]$.
\begin{equation}
\begin{aligned}
&\mathbb{E}[\|\nabla F(\mathbf{x}_{t+r})-W'\mathbf{g}(\mathbf{x}_t)\|]
\leq\mathbb{E}[\|\nabla F(\mathbf{x}_{t+r})-\mathbf{g}(\mathbf{x}_t)\|]\\
=&\mathbb{E}[\|\nabla F(\mathbf{x}_{t+r})-\nabla F(\mathbf{x}_t)+\nabla F(\mathbf{x}_t)-\mathbf{g}(\mathbf{x}_t)\|]\\
\leq& \mathbb{E}[\|\nabla F(\mathbf{x}_{t+r})-\nabla F(\mathbf{x}_t)\|]+\mathbb{E}[\|\nabla F(\mathbf{x}_t)-\mathbf{g}(\mathbf{x}_t)\|]\\
\leq&\gamma_m\|\mathbf{x}_{t+r}-\mathbf{x}_t\|+\sqrt{(\mathbb{E}[\|\nabla F(\mathbf{x}_t)-\mathbf{g}(\mathbf{x}_t)\|])^{2}}\\
\leq& \gamma_m\eta\frac{2G+(r-1)B\theta_m}{1-\delta_2}+\sqrt{\mathbb{E}[\|\nabla F(\mathbf{x}_t)-\mathbf{g}(\mathbf{x}_t)\|]^{2}}\\
\leq& \gamma_m\eta\frac{2G+(r-1)B\theta_m}{1-\delta_2}+\sigma
\end{aligned}
\end{equation}
Hence, we have 
\begin{equation}
\begin{aligned}
&\|\mathbb{E}[\mathcal{G}^h(\mathbf{x}_t)]-\sum_{r=0}^{\tau-1}W'\mathbf{g}^{dc,r}(\mathbf{x}_t)\|\\
\leq& \gamma_m\eta\frac{2G+(r-1)B\theta_m}{1-\delta_2}+[(1-\lambda)G^2
+\epsilon_D+\epsilon]\\&
\eta\frac{2G+(r-1)B\theta_m}{1-\delta_2}
+\sigma\\
=&(\gamma_m+\epsilon_D+\epsilon+(1-\lambda)G^2)\eta\frac{2G+(r-1)B\theta_m}{1-\delta_2}+\sigma
\end{aligned}
\end{equation}

The above relationship is obtained for cases where $r\geq1$. There still is $r=0$ left.
For $r=0$,
\begin{equation}
\|\nabla F(\mathbf{x}_t)-W'\mathbf{g}(\mathbf{x}_t)\|\leq\sigma
\end{equation}
Thus, combining each upper bound for $\|\mathbb{E}[\mathcal{G}^{h,r}(\mathbf{x}_t)]-W'\mathbf{g}^{dc,r}(\mathbf{x}_t)\|$,
we can obtain
\begin{equation}
\begin{aligned}
&\|\mathbb{E}[\mathcal{G}^h(\mathbf{x}_t)]-\sum_{r=0}^{\tau-1}W'\mathbf{g}^{dc,r}(\mathbf{x}_t)\|
\\&\leq\sum_{r=1}^{\tau-1}(\gamma_m+\epsilon_D+\epsilon+(1-\lambda)G^2)\eta\frac{2G+(r-1)B\theta_m}{1-\delta_2}+\tau\sigma,
\end{aligned}
\end{equation}

which completes the proof.
\end{proof}

\noindent $\textbf{Lemma 5}$:
Let Assumptions 1, 2 and 3 hold. Assume that the delay compensated gradients are uniformly bounded, i.e., there exists a scalar $B>0$ such that 
\begin{equation}
   \|\mathbf{g}^{dc,r}(\mathbf{x}_{t})\|\leq B,\,\,\ \forall t\geq 0 \,\ and \,\ 0\leq r\leq\tau-1,
\end{equation}

Then for the iterates generated by PC-ASGD, $\exists \eta>0$, they satisfy
\begin{equation}
\begin{aligned}
&F(\mathbf{x}_{t+\tau})\geq F(\mathbf{v}_{t+\tau})-2G\eta\frac{G+(\tau-1)B\theta_m}{1-\delta_2}
\end{aligned}
\end{equation}

\begin{proof}
Due to the convexity, we have 
\begin{equation}
\begin{aligned}
F(\mathbf{x}_{t+\tau})&\geq F(\mathbf{v}_{t+\tau})+\nabla F(\mathbf{v}_{t+\tau})(\mathbf{x}_{t+\tau}-\mathbf{v}_{t+\tau})\\&\geq F(\mathbf{v}_{t+\tau})-\|\nabla F(\mathbf{v}_{t+\tau})\|\|\mathbf{v}_{t+\tau}-\mathbf{x}_{t+\tau}\|\\
&\geq F(\mathbf{v}_{t+\tau})-G\|\mathbf{v}_{t+\tau}-\mathbf{x}_{t+\tau}\|\\&\geq F(\mathbf{v}_{t+\tau})-G\|\mathbf{v}_{t+\tau}-y_{t+\tau}\mathbf{1}+y_{t+\tau}\mathbf{1}-\mathbf{x}_{t+\tau}\|\\
&\geq F(\mathbf{v}_{t+\tau})-G(\|\mathbf{v}_{t+\tau}-y_{t+\tau}\mathbf{1}\|+\|y_{t+\tau}\mathbf{1}-\mathbf{x}_{t+\tau}\|)\\
&\geq F(\mathbf{v}_{t+\tau})-2G\eta\frac{G+(\tau-1)B\theta_m}{1-\delta_2}
\end{aligned}
\end{equation}

The second inequality follows from the Cauthy-Schwarz inequality. The proof is completed.
\end{proof}
\noindent $\textbf{Theorem~\ref{PL_theorem}}$:
Let Assumptions 1,2 and 3 hold. Assume that the delay compensated gradients are uniformly bounded, i.e., there exits a a scalar $B>0$ such that
\begin{equation}
    \|\mathbf{g}^{dc,r}(\mathbf{x}_{t})\|\leq B,\,\,\ \forall t\geq 0 \,\ and \,\ 0\leq r\leq\tau-1,
\end{equation}
and that $\nabla F(\mathbf{x}_t)$ is $\xi_m$-smooth for all $t\geq 0$. Then for the iterates generated by PC-ASGD, when $0<\eta\leq\frac{1}{2\mu\tau}$ and the objective satisfies the PL condition, they satisfy
\begin{equation}
\begin{split}
\mathbb{E}[F(\mathbf{x}_{t})-F^{*}]&\leq (1-2\mu\eta\tau)^{t-1}(F(\mathbf{x}_1)-F^{*}-\frac{Q}{2\mu\eta\tau})\\&+\frac{Q}{2\mu\eta\tau},
\end{split}
\end{equation}
\begin{equation}
{
\begin{aligned}
    Q &= 2(1-2\mu\eta\tau)G\eta C_1+\frac{\eta^3\xi_mG}{2}\sum_{r=1}^{\tau-1}C_r
+2\eta^2G\gamma_m C_1\\[-6pt]
&+G\eta\tau\sigma+\eta^2G(\gamma_m+\epsilon_D+\epsilon+(1-\lambda)G^2)\sum_{r=1}^{\tau-1}C_r\\[-6pt]
&+\eta G^2+\eta^2\gamma_mG\tau C_2\\
\end{aligned}
}
\end{equation}
and,
\begin{equation}
{
\begin{aligned}
&C_1=\frac{G+(\tau-1)B\theta_m}{1-\delta_2}\\
&C_r=\frac{2G+(r-1)B\theta_m}{1-\delta_2}\\
&C_2=\frac{2G+(\tau-1)B\theta_m}{1-\delta_2},\\
\end{aligned}
}
\end{equation}
$\epsilon_D > 0$ and $\epsilon > 0$ are upper bounds for the approximation errors of the Hessian matrix.
\begin{proof}
According to the smoothness condition of $F(\mathbf{x})$. We have
\begin{equation}
\begin{aligned}
&\mathbb{E}[F(\mathbf{x}_{t+\tau+1})-F(\mathbf{x}^{*})]\leq\mathbb{E}[F(\mathbf{v}_{t+\tau})-F(\mathbf{x}^{*})]\\&+\mathbb{E}[\langle\nabla F(\mathbf{v}_{t+\tau}),(\mathbf{x}_{t+\tau+1}-\mathbf{v}_{t+\tau})\rangle]\\&+\frac{\gamma_m}{2}\mathbb{E}[\|\mathbf{x}_{t+\tau+1}-\mathbf{v}_{t+\tau}\|^2]\\
\end{aligned}
\end{equation}

Based on the update law, we can obtain 
\begin{equation}
\begin{aligned}
&\mathbb{E}[F(\mathbf{x}_{t+\tau+1})-F(\mathbf{x}^{*})]
\\&\leq\mathbb{E}[F(\mathbf{v}_{t+\tau})-F^{*}]-\eta\mathbb{E}[\langle\nabla F(\mathbf{v}_{t+\tau}),\mathbf{g}(\mathbf{x}_{t+\tau})\rangle]-\\
&\eta\mathbb{E}[\langle\nabla F(\mathbf{v}_{t+\tau}),\sum_{r=0}^{\tau-1}W'\mathbf{g}^{dc,r}(\mathbf{x}_t)\rangle]\\&+\frac{\gamma_m\eta^2}{2}\mathbb{E}[\|\mathbf{g}(\mathbf{x}_{t+\tau})+\sum_{r=0}^{\tau-1}W'\mathbf{g}^{dc,r}(\mathbf{x}_t)\|^2] \\
&\leq\mathbb{E}[F(\mathbf{v}_{t+\tau})-F^{*}]-\eta\mathbb{E}[\langle\nabla F(\mathbf{v}_{t+\tau}),\mathbf{g}(\mathbf{x}_{t+\tau})\rangle]\\&-\eta\mathbb{E}[\langle\nabla F(\mathbf{v}_{t+\tau}),\tau\nabla F(\mathbf{v}_{t+\tau})\rangle] \\
&+\eta\mathbb{E}[\langle\nabla F(\mathbf{v}_{t+\tau}),\tau\nabla F(\mathbf{v}_{t+\tau})-\sum_{r=0}^{\tau-1}\nabla F(\mathbf{v}_{t+r})\rangle]\\&+\eta\mathbb{E}[\langle\nabla F(\mathbf{v}_{t+\tau}),\sum_{r=0}^{\tau-1}\nabla F(\mathbf{v}_{t+r})-\mathcal{F}^{h}(\mathbf{x}_t)\rangle]\\
&+\eta\mathbb{E}[\langle\nabla F(\mathbf{v}_{t+\tau}),\mathbb{E}[\mathcal{G}^{h}]-\sum_{r=0}^{\tau-1}W'\mathbf{g}^{dc,r}(\mathbf{x}_t)\rangle]\\&+\frac{\gamma_m\eta^2}{2}\mathbb{E}[\|\mathbf{g}(\mathbf{x}_{t+\tau})
+\sum_{r=0}^{\tau-1}W'\mathbf{g}^{dc,r}(\mathbf{x}_t)\|^2]
\end{aligned}
\end{equation}

We next investigate each term on the right hand side. Based on Lemma 4, we can obtain
\begin{equation}
\begin{aligned}
F(\mathbf{x}_{t+\tau})\geq F(\mathbf{v}_{t+\tau})-2G\eta\frac{G+(\tau-1)B\theta_m}{1-\delta_2}
\end{aligned}
\end{equation}

such that
\begin{equation}
\begin{aligned}
F(\mathbf{x}_{t+\tau})-F^{*}
\geq F(\mathbf{v}_{t+\tau})-F^{*}-2G\eta\frac{G+(\tau-1)B\theta_m}{1-\delta_2}
\end{aligned}
\end{equation}

For the term $-\eta\mathbb{E}[\langle\nabla F(\mathbf{v}_{t+\tau}),g(\mathbf{x}_{t+\tau})\rangle]$, we can quickly get that is is bounded above by $\eta G^2$ due to the Cauthy-Schwarz inequality.
Then for term $-\eta\mathbb{E}[\langle\nabla F(\mathbf{v}_{t+\tau}),\tau\nabla F(\mathbf{v}_{t+\tau})\rangle]$, one can get the following relationship due to the PL condition.
\begin{equation}
\begin{aligned}
-\eta\mathbb{E}&[\langle \nabla F(\mathbf{v}_{t+\tau}),\tau\nabla F(\mathbf{v}_{t+\tau})\rangle]
\leq -2\eta\tau\mu(F(\mathbf{v}_{t+\tau})-F^{*})
\end{aligned}
\end{equation}

combining $F(\mathbf{v}_{t+\tau})-F^{*}$, we have
\begin{equation}
\begin{aligned}
&(1-2\eta\tau\mu)(F(\mathbf{v}_{t+\tau})-F^{*})
\\&\leq (1-2\eta\tau\mu)[(F(\mathbf{x}_{t+\tau})-F^{*})
+2G\eta\frac{G+(\tau-1)B\theta_m}{1-\delta_2}]
\end{aligned}
\end{equation}

Based on Lemma 2, we have known that
\begin{equation}
\begin{aligned}
&\|\nabla F(\mathbf{v}_{t+r})-\nabla \mathcal{F}^{h,r}(\mathbf{x}_t)\|\leq\frac{\xi_m}{2}\eta^2[\frac{2G+(r-1)B\theta_m}{1-\delta_2}]^{2};
\end{aligned}
\end{equation}

for $r\geq1$, while for $r=0$, it can be obtained that
\begin{equation}
\begin{aligned}
&\|\nabla F(\mathbf{v}_{t})-\nabla F(\mathbf{x}_t))\|
\leq 2\gamma_m\eta\frac{G+(\tau-1)B\theta_m}{1-\delta_2}.
\end{aligned}
\end{equation}

Since
\begin{equation}
\begin{aligned}
&\eta\mathbb{E}[\langle \nabla F(\mathbf{v}_{t+\tau}),\sum_{r=0}^{\tau-1}\nabla F(\mathbf{v}_{t+r})-\mathcal{F}^{h}(\mathbf{x}_t)\rangle]\\
&\leq\eta\mathbb{E}[\|\nabla F(\mathbf{v}_{t+\tau})\|\|\sum_{r=0}^{\tau-1}\nabla F(\mathbf{v}_{t+r})-\mathcal{F}^{h}(\mathbf{x}_t)\|]\\
&\leq \mathbb{E}[\|\nabla F(\mathbf{v}_{t+\tau})\|\sum_{r=0}^{\tau-1}\|\nabla F(\mathbf{v}_{t+r})-\mathcal{F}^{h}(\mathbf{x}_t)\|]\\
\end{aligned}
\end{equation}

The first inequality follows from Cauthy-Schwarz inequality and the second one follows from Triangle inequality. Hence, we can have
\begin{equation}
\begin{aligned}
\eta\mathbb{E}[\langle\nabla &F(\mathbf{v}_{t+\tau}),\sum_{r=0}^{\tau-1}\nabla F(\mathbf{v}_{t+r})-\mathcal{F}^{h}(\mathbf{x}_t)\rangle]
\\&\leq \frac{\eta^3\xi_mG}{2(1-\delta_2)}\sum_{r=1}^{\tau-1}[2G+B(r-1)\theta_m]\\
&+2\eta^2G\gamma_m\frac{G+(\tau-1)B\theta_m}{1-\delta_2}
\end{aligned}
\end{equation}

According to Lemma 3, the following relationship can be obtained,
\begin{equation}
\begin{aligned}
\mathbb{E}[\langle\nabla &F(\mathbf{v}_{t+\tau}),\mathbb{E}[\mathcal{G}^h(\mathbf{x}_t)]-\sum_{r=0}^{\tau-1}W'\mathbf{g}^{dc,r}(\mathbf{x}_t)\rangle]\\&\leq\frac{\eta^2G}{1-\delta_2}(\gamma_m+\epsilon_D+\epsilon+(1-\lambda)G^2)\\&\sum_{r=1}^{\tau-1}[2G+(r-1)B\theta_m]
+G\eta\tau\sigma
\end{aligned}
\end{equation}

The last term is $\eta\mathbb{E}[\langle\nabla F(\mathbf{v}_{t+\tau}),\tau\nabla F(\mathbf{v}_{t+\tau})-\sum_{r=0}^{\tau-1}\nabla F(\mathbf{v}_{t+r})\rangle]$, which can be rewritten such that
\begin{equation}
\begin{aligned}
&\eta\mathbb{E}[\langle\nabla F(\mathbf{v}_{t+\tau}),\tau\nabla F(\mathbf{v}_{t+\tau})-\sum_{r=0}^{\tau-1}\nabla F(\mathbf{v}_{t+r})\rangle]\\&\leq\eta\mathbb{E}[\|\nabla F(\mathbf{v}_{t+\tau})\|\|\nabla F(\mathbf{v}_{t+\tau})-\nabla F(\mathbf{v}_{t})
\\&+...+\nabla F(\mathbf{v}_{t+\tau})-\nabla F(\mathbf{v}_{t+\tau-1})\|]\\
&\leq\eta\mathbb{E}[\|\nabla F(\mathbf{v}_{t+\tau})\|\|\nabla F(\mathbf{v}_{t+\tau})-\nabla F(\mathbf{v}_{t})\|\\&+...+\|\nabla F(\mathbf{v}_{t+\tau})-\nabla F(\mathbf{x}_{t+\tau-1})\|]
\end{aligned}
\end{equation}

Using the smoothness condition, we then can bound the term by deriving the following relationship with Lemma 1 and Lemma 2, 
\begin{equation}
\begin{aligned}
&\eta\mathbb{E}[\langle\nabla F(\mathbf{v}_{t+\tau}),\tau\nabla F(\mathbf{v}_{t+\tau})-\sum_{r=0}^{\tau-1}\nabla F(\mathbf{v}_{t+r})\rangle]\\&\leq \eta^2\gamma_mG\tau\frac{2G+(\tau-1)B\theta_m}{1-\delta_2}
\end{aligned}
\end{equation}

We combine the upper bounds of each term on the right hand side to produce the following relationship.
\begin{equation}
\begin{aligned}
&\mathbb{E}[F(\mathbf{x}_{t+\tau+1})-F(\mathbf{x}^{*})\\
&\leq(1-2\eta\mu\tau)(F(\mathbf{x}_{t+\tau})-F^{*})\\&+2(1-2\eta\mu\tau)G \eta\frac{G+(\tau-1)B\theta_m}{1-\delta_2}\\
&+\frac{\eta^3\xi_mG}{2(1-\delta_2)}\sum_{r=1}^{\tau-1}[2G+(r-1)B\theta_m]\\&+2\eta^2G\gamma_m\frac{G+(\tau-1)B\theta_m}{1-\delta_2}\\
&+G\eta\tau\sigma+\frac{\eta^2G}{1-\delta_2}(\gamma_m+\epsilon_D+\epsilon
\\&+(1-\lambda)G^2)\sum_{r=1}^{\tau-1}[2G+(r-1)B\theta_m]\\&+\eta^2\gamma_mG\tau\frac{2G+(\tau-1)B\theta_m}{1-\delta_2}
+\eta G^2.
\end{aligned}
\end{equation}




We have now known that 
\begin{equation}
\begin{aligned}
    \mathbb{E}[F(\mathbf{x}_{t+1})-F^{*}]\leq (1-2\eta\tau\mu)\mathbb{E}[F(\mathbf{x}_t)-F^{*}]+Q,
\end{aligned}
\end{equation}

subtracting the constant $\frac{Q}{2\mu\tau\eta}$ from both sides, one obtains 
\begin{equation}
\begin{aligned}
    \mathbb{E}[F(\mathbf{x}_{t+1})-F^{*}]-\frac{Q}{2\mu\tau\eta}&\leq (1-2\eta\mu\tau)\mathbb{E}[F(\mathbf{x}_t)-F^{*}]\\
    &+Q-\frac{Q}{2\mu\tau\eta}\\&=(1-2\eta\mu\tau)(\mathbb{E}[F(\mathbf{x}_t)-F^{*}]\\&-\frac{Q}{2\mu\tau\eta})
\end{aligned}
\end{equation}

Observe that the above inequality is a contraction inequality since $0<2\eta\mu\tau\leq 1$ due to $0<\eta\leq\frac{1}{2\mu\tau}$. The result thus follows by applying the inequality repeatedly through iteration $t\in\mathbb{N}$.
\end{proof}
Another scenario that could be of interest is the strongly convex objective. As Theorem~\ref{PL_theorem} has shown that with a properly set constant step size, PC-ASGD enables to converge to the neighborhood of the optimal solution with a linear rate. This also applies to the strongly convex objective in which the strong convexity implies the PL condition, while the constants are subject to changes. We now proceed to give the proof for the nonconvex case.

\noindent $\textbf{Theorem~\ref{Non_convex}}$: Let Assumptions 1,2 and 3 hold. Assume that the delay compensated gradients are uniformly bounded, i.e., there exits a a scalar $B>0$ such that 
\begin{equation}
    \|\mathbf{g}^{dc,r}(\mathbf{x}_{t})\|\leq B,\,\,\ \forall t\geq 0 \,\ and \,\ 0\leq r\leq\tau-1,
\end{equation}
and that
\begin{equation}
    \mathbb{E}[\|\mathbf{g}^{dc,r}(\mathbf{x}_t)\|^2]\leq M.
\end{equation}

Then for the iterates generated by PC-ASGD, there exists $0<\eta<\frac{1}{\gamma_m}$, such that for all $T\geq 1$,
\begin{equation}
\begin{aligned}
\frac{1}{T}\sum_{t=1}^{T}\mathbb{E}[\|\nabla F(\mathbf{x}_t)\|^2]\leq \frac{2(F(\mathbf{x}_1)-F^{*})}{T\eta}+\frac{R}{\eta},
\end{aligned}
\end{equation}
where\[
R=2G\eta^2 C_1+\frac{\tau\eta^2\gamma_mM}{2}+\frac{\eta\sigma^2}{2}+\eta\sigma\tau B+2\eta^2\gamma_m(\tau B+G)C_1.\]

\begin{proof}
According to the smoothness condition of $F(\mathbf{x})$, we have
\begin{align*}
&F(\mathbf{x}_{t+\tau+1})-F(\mathbf{v}_{t+\tau})\\
&\leq \langle\nabla F(\mathbf{v}_{t+\tau}), \mathbf{x}_{t+\tau+1}-\mathbf{v}_{t+\tau}\rangle+\frac{\gamma_m}{2}+\|\mathbf{x}_{t+\tau+1}-\mathbf{v}_{t+\tau}\|^{2}\\
&=\langle\nabla F(\mathbf{v}_{t+\tau}),-\eta(\sum_{r=0}^{\tau-1}W'\mathbf{g}^{dc,r}(\mathbf{x}_t)+\mathbf{g}(\mathbf{x}_{t+\tau}))\rangle\\&+\frac{\eta^2\gamma_m}{2}\|\sum_{r=0}^{\tau-1}W'\mathbf{g}^{dc,r}+\mathbf{g}(\mathbf{x}_{t+\tau})\|^2\\
&=\langle\nabla F(\mathbf{v}_{t+\tau})-\nabla F(\mathbf{x}_{t+\tau})+\nabla F(\mathbf{x}_{t+\tau}),\\&\eta(\sum_{r=0}^{\tau-1}W'\mathbf{g}^{dc,r}(\mathbf{x}_t)+\mathbf{g}(\mathbf{x}_{t+\tau}))\rangle\\&+\frac{\eta^2\gamma_m}{2}\|\sum_{r=0}^{\tau-1}W'\mathbf{g}^{dc,r}+\mathbf{g}(\mathbf{x}_{t+\tau})\|^2\\
&=-\eta\langle\nabla F(\mathbf{x}_{t+\tau}),\sum_{r=0}^{\tau-1}W'\mathbf{g}^{dc,r}(\mathbf{x}_t)+\mathbf{g}(\mathbf{x}_{t+\tau})\rangle\\&+\eta\langle(\nabla F(\mathbf{v}_{t+\tau})-\nabla F(\mathbf{x}_{t+\tau}),\sum_{r=0}^{\tau-1}W'\mathbf{g}^{dc,r}(\mathbf{x}_t)+\mathbf{g}(\mathbf{x}_{t+\tau}))\rangle\\&+\frac{\eta^2\gamma_m}{2}\|\sum_{r=0}^{\tau-1}W'\mathbf{g}^{dc,r}+\mathbf{g}(\mathbf{x}_{t+\tau})\|^2\\
&=-\frac{\eta}{2}[\|\nabla F(\mathbf{x}_{t+\tau})\|^2+\|\sum_{r=0}^{\tau-1}W'\mathbf{g}^{dc,r}(\mathbf{x}_t)+\mathbf{g}(\mathbf{x}_{t+\tau})\|^2\\&-\|\nabla F(\mathbf{x}_{t+\tau})-(\sum_{r=0}^{\tau-1}W'\mathbf{g}^{dc,r}(\mathbf{x}_t)+\mathbf{g}(\mathbf{x}_{t+\tau}))\|^2]\\
&+\eta\langle\nabla F(\mathbf{x}_{t+\tau})-\nabla F(\mathbf{v}_{t+\tau}),\sum_{r=0}^{\tau-1}W'\mathbf{g}^{dc,r}(\mathbf{x}_t)+\mathbf{g}(\mathbf{x}_{t+\tau})\rangle\\&+\frac{\eta^2\gamma_m}{2}\|\sum_{r=0}^{\tau-1}W'\mathbf{g}^{dc,r}+\mathbf{g}(\mathbf{x}_{t+\tau})\|^2\\
&=-\frac{\eta}{2}\|\nabla F(\mathbf{x}_{t+\tau})\|^2-\frac{\eta}{2}\|\sum_{r=0}^{\tau-1}W'\mathbf{g}^{dc,r}(\mathbf{x}_t)+\mathbf{g}(\mathbf{x}_{t+\tau})\|^2\\&+\frac{\eta}{2}(\|\nabla F(\mathbf{x}_{t+\tau})-\mathbf{g}(\mathbf{x}_{t+\tau})\|^2+\|\sum_{r=0}^{\tau-1}W'\mathbf{g}^{dc,r}(\mathbf{x}_t)\|^2\\&-2\langle\nabla F(\mathbf{x}_{t+\tau})-\mathbf{g}(\mathbf{x}_{t+\tau}),\sum_{r=0}^{\tau-1}W'\mathbf{g}^{dc,r}(\mathbf{x}_t)\rangle)\\&+\eta\langle\nabla F(\mathbf{x}_{t+\tau})-\nabla F(\mathbf{v}_{t+\tau}),\sum_{r=0}^{\tau-1}W'\mathbf{g}^{dc,r}(\mathbf{x}_t)+\mathbf{g}(\mathbf{x}_{t+\tau})\rangle\\&+\frac{\eta^2\gamma_m}{2}\|\sum_{r=0}^{\tau-1}W'\mathbf{g}^{dc,r}+\mathbf{g}(\mathbf{x}_{t+\tau})\|^2\\ 
&=-\frac{\eta}{2}\|\nabla F(\mathbf{x}_{t+\tau})\|^2-(\frac{\eta}{2}-\frac{\eta^2\gamma_m}{2})\|\sum_{r=0}^{\tau-1}W'\mathbf{g}^{dc,r}(\mathbf{x}_t)  \notag+\mathbf{g}(\mathbf{x}_{t+\tau})\|^2\\&+\frac{\eta}{2}\|\nabla F(\mathbf{x}_{t+\tau})-\mathbf{g}(\mathbf{x}_{t+\tau})\|^2 \notag+\frac{\eta}{2}\|\sum_{r=1}^{\tau-1}W'\mathbf{g}^{dc,r}(\mathbf{x}_t)\|^2\notag\\&-\eta\langle\nabla F(\mathbf{x}_{t+\tau})-\mathbf{g}(\mathbf{x}_{t+\tau}),\sum_{r=1}^{\tau-1}W'\mathbf{g}^{dc,r}(\mathbf{x}_t)\rangle  \notag\\&+\eta\langle\nabla F(\mathbf{x}_{t+\tau})-\nabla F(\mathbf{v}_{t+\tau}),\sum_{r=0}^{\tau-1}W'\mathbf{g}^{dc,r}(\mathbf{x}_t)+\mathbf{g}(\mathbf{x}_{t+\tau})\rangle  \notag\\
&=-\frac{\eta}{2}\|\nabla F(\mathbf{x}_{t+\tau})\|^2+(\frac{\eta^2\gamma_m}{2}-\frac{\eta}{2})\|\sum_{r=0}^{\tau-1}W'\mathbf{g}^{dc,r}(\mathbf{x}_t)\|^2 \notag\\&+(\frac{\eta^2\gamma_m}{2}-\frac{\eta}{2})\|\mathbf{g}(\mathbf{x}_{t+\tau})\|^2\\& \notag\\&+(\frac{\eta^2\gamma_m}{2}-\frac{\eta}{2})\langle\mathbf{g}(\mathbf{x}_{t+\tau}),\sum_{r=0}^{\tau-1}W'\mathbf{g}^{dc,r}(\mathbf{x}_t)\rangle \notag\\
&+\frac{\eta}{2}\|\nabla F(\mathbf{x}_{t+\tau})-\mathbf{g}(\mathbf{x}_{t+\tau})\|^2+\frac{\eta}{2}\|\sum_{r=1}^{\tau-1}W'\mathbf{g}^{dc,r}(\mathbf{x}_t)\|^2  \notag\\&-\eta\langle\nabla F(\mathbf{x}_{t+\tau})-\mathbf{g}(\mathbf{x}_{t+\tau}),\sum_{r=1}^{\tau-1}W'\mathbf{g}^{dc,r}(\mathbf{x}_t)\rangle  \notag\\
&+\eta\langle\nabla F(\mathbf{x}_{t+\tau})-\nabla F(\mathbf{v}_{t+\tau}),\sum_{r=0}^{\tau-1}W'\mathbf{g}^{dc,r}(\mathbf{x}_t)+\mathbf{g}(\mathbf{x}_{t+\tau})\rangle  \notag\\
&\leq -\frac{\eta}{2}\|\nabla F(\mathbf{x}_{t+\tau})\|^2+(\frac{\eta^2\gamma_m}{2}-\frac{\eta}{2})\|\sum_{r=0}^{\tau-1}W'\mathbf{g}^{dc,r}(\mathbf{x}_t)\|^2 \notag\\&+(\frac{\eta^2\gamma_m}{2}-\frac{\eta}{2})\|\mathbf{g}(\mathbf{x}_{t+\tau})\|^2\notag\\&+(\frac{\eta^2\gamma_m}{2}-\frac{\eta}{2})\|\mathbf{g}(\mathbf{x}_{t+\tau})\|\|\sum_{r=0}^{\tau-1}W'\mathbf{g}^{dc,r}(\mathbf{x}_t)\| \notag\\
&+\frac{\eta}{2}\|\nabla F(\mathbf{x}_{t+\tau})-\mathbf{g}(\mathbf{x}_{t+\tau})\|^2+\frac{\eta}{2}\|\sum_{r=1}^{\tau-1}W'\mathbf{g}^{dc,r}(\mathbf{x}_t)\|^2  \notag\\
&+\eta\|\nabla F(\mathbf{x}_{t+\tau})-\mathbf{g}(\mathbf{x}_{t+\tau})\|\|\sum_{r=1}^{\tau-1}W'\mathbf{g}^{dc,r}(\mathbf{x}_t)\|  \notag\\&+\eta\|\nabla F(\mathbf{x}_{t+\tau})-\nabla F(\mathbf{v}_{t+\tau})\| \notag \|\sum_{r=0}^{\tau-1}W'\mathbf{g}^{dc,r}(\mathbf{x}_t)+\mathbf{g}(\mathbf{x}_{t+\tau})\|.
\end{align*}

The first inequality follows from the smooth property of the objective. The last inequality follows from Cauthy-Schwarz inequality.

The left hand side of the above inequality can be rewritten associated
\begin{align}
F(\mathbf{x}_{t+\tau+1})-F(\mathbf{x}_{t+\tau})+F(\mathbf{x}_{t+\tau})-F(\mathbf{v}_{t+\tau}) \notag
\end{align}

Taking expectation for both sides, with the last inequality, we have 
\begin{align*}
&\mathbb{E}[F({\mathbf{x}_{t+\tau+1})}-F(\mathbf{x}_{t+\tau})]\\
\leq&\mathbb{E}[F(\mathbf{v}_{t+\tau})-F(\mathbf{x}_{t+\tau})]-\frac{\eta}{2}\mathbb{E}[\|\nabla F(\mathbf{x}_{t+\tau})\|^2]\\
&+\frac{\eta^2\gamma_m-\eta}{2}\mathbb{E}[\|\sum_{r=0}^{\tau-1}W'\mathbf{g}^{dc,r}(\mathbf{x}_t)\|^2]\\&+\frac{\eta^2\gamma_m-\eta}{2}\mathbb{E}[\|\mathbf{g}(\mathbf{x}_{t+\tau})\|^2]
\\&+\frac{\eta^2\gamma_m-\eta}{2}\mathbb{E}[\|\mathbf{g}(\mathbf{x}_{t+\tau})\|\|\sum_{r=0}^{\tau-1}W'\mathbf{g}^{dc,r}(\mathbf{x}_t)\|]\\&+\frac{\eta}{2}\mathbb{E}[\|\nabla F(\mathbf{x}_{t+\tau})-\mathbf{g}(\mathbf{x}_{t+\tau})\|^2]\\&+\frac{\eta}{2}\mathbb{E}[\|\sum_{r=1}^{\tau-1}W'\mathbf{g}^{dc,r}(\mathbf{x}_t)\|^2]\\&+\eta\mathbb{E}[\|\nabla F(\mathbf{x}_{t+\tau})-\mathbf{g}(\mathbf{x}_{t+\tau})\|\|\sum_{r=1}^{\tau-1}W'\mathbf{g}^{dc,r}(\mathbf{x}_t)\|]\\&+\eta\mathbb{E}[\|\nabla F(\mathbf{x}_{t+\tau})-\nabla F(\mathbf{v}_{t+\tau})\|\\&\|\sum_{r=0}^{\tau-1}W'\mathbf{g}^{dc,r}(\mathbf{x}_t)+\mathbf{g}(\mathbf{x}_{t+\tau})\|]\\
\leq& G\mathbb{E}[\|\mathbf{v}_{t+\tau}-\mathbf{x}_{t+\tau}\|]-\frac{\eta}{2}\mathbb{E}[\|\nabla F(\mathbf{x}_{t+\tau})\|^2]\\
&+\frac{\eta^2\gamma_m-\eta}{2}\tau\sum_{r=0}^{\tau-1}\mathbb{E}[\|W'\mathbf{g}^{dc,r}(\mathbf{x}_t)\|^2]\\&+\frac{\eta^2\gamma_m-\eta}{2}\mathbb{E}[\|\mathbf{g}(\mathbf{x}_{t+\tau})\|^2]\\
&+\frac{\eta^2\gamma_m-\eta}{2}\mathbb{E}[\|\mathbf{g}(\mathbf{x}_{t+\tau})\|\|\sum_{r=0}^{\tau-1}W'\mathbf{g}^{dc,r}(\mathbf{x}_t)\|]\\&+\frac{\eta}{2}\mathbb{E}[\|\nabla F(\mathbf{x}_{t+\tau})-\mathbf{g}(\mathbf{x}_{t+\tau})\|^2]\\&+\frac{\eta}{2}\mathbb{E}[\|\sum_{r=1}^{\tau-1}W'\mathbf{g}^{dc,r}(\mathbf{x}_t)\|^2]\\
&+\eta\mathbb{E}[\|\nabla F(\mathbf{x}_{t+\tau})-\mathbf{g}(\mathbf{x}_{t+\tau})\|\|\sum_{r=1}^{\tau-1}W'\mathbf{g}^{dc,r}(\mathbf{x}_t)\|]\\&+\eta\mathbb{E}[\|\nabla F(\mathbf{x}_{t+\tau})-\nabla F(\mathbf{v}_{t+\tau})\|\\&\|\sum_{r=0}^{\tau-1}W'\mathbf{g}^{dc,r}(\mathbf{x}_t)+\mathbf{g}(\mathbf{x}_{t+\tau})\|]\\
\leq& -\frac{\eta}{2}\mathbb{E}[\|\nabla F(\mathbf{x}_{t+\tau})\|^2]\\&+\frac{\tau^2\eta^2\gamma_mM}{2}+\frac{\eta\sigma^2}{2}+\eta\sigma\tau B\\
&+2\eta^2\gamma_m(\tau B+G+\frac{G}{\eta\gamma_m})\frac{G+(\tau-1)B\theta_m}{1-\delta_2}
\end{align*}
The last inequality follows from the smoothness condition of $F(\mathbf{x})$ and the bounded gradient, respectively, as well as $\eta<\frac{1}{\gamma_m}$. Hence, by replacing $t+\tau$ with $t$, one can obtain
\begin{equation}
\begin{aligned}
\mathbb{E}[F(\mathbf{x}_{t+1})-F(\mathbf{x}_t)]\leq-\frac{\eta}{2}\mathbb{E}[\|\nabla F(\mathbf{x}_t)\|^2]+R
\end{aligned}
\end{equation}

where $R$ indicates the constant term on the right hand side of the inequality.
As we assume that $F(\mathbf{x})$ is bounded from below, applying the last inequality from 1 to $T$, one can get 
\begin{equation}
\begin{aligned}
F^{*}-F(\mathbf{x}_1)&\leq \mathbb{E}[F(\mathbf{x}_{t+1})]-F(\mathbf{x}_1)\\
&\leq-\frac{\eta}{2}\sum_{t=1}^{T}\mathbb{E}[\|\nabla F(\mathbf{x}_t)\|^2]+TR
\end{aligned}
\end{equation}

which results in
\begin{equation}
\begin{aligned}
\sum_{t=1}^{T}\mathbb{E}[\|\nabla F(\mathbf{x}_t)\|^2]\leq\frac{2[(F(\mathbf{x}_1)-F^{*})+TR]}{\eta}
\end{aligned}
\end{equation}

Dividing both sides by $T$, the desirable results is obtained.
\end{proof}

\textbf{Detailed Settings of Deep Learning Models.}
For the PreResNet110 (\emph{model 1}) and DenseNet (\emph{model 2}), the batch size is selected as 128. After hyperparameter searching in $(0.1, 0.01, 0.001)$, the learning rate is set as 0.01 for the first 160 epochs and changed as 0.001 . The decay are applied in epochs $(80,120,160,200)$. The approximation coefficient $\lambda$ is set as 1. $\lambda=0.001$ is first tried as suggested by DC-ASGD \cite{zheng2017asynchronous} and the results show that the predicting step doesn't affect the training process. By considering the upper bound of 1, a set of values $(0.001,0.1,1)$ are tried and $\lambda=1$ is applied according to the performance.

As for the practical implementation, an structure that is much closer to the real distributed system is used. Each agent is allocated to an independent GPU, and a communication layer is set up for the parameter transferring, which is convenient to the following protocol design. Such settings provide us availability for quick implementation of the algorithm in real distributed networks.

\end{document}